\documentclass[11pt]{article}

\usepackage{emnlp2022}

\usepackage{times}
\usepackage{latexsym}
\usepackage{amsmath}
\usepackage{mathtools}
\usepackage{amssymb}
\usepackage{tikz}
\usepackage{tikz-qtree}
\usepackage{paralist, tabularx} 

\usepackage{booktabs} 
\usepackage{multicol}
\usepackage{multirow}

\definecolor{Gray}{gray}{0.9}
\usepackage{soul}
\DeclareRobustCommand{\hlgray}[1]{{\sethlcolor{Gray}\hl{#1}}}

\usepackage{color}

\newcommand{\affa}{{$^{1}$}}
\newcommand{\affb}{{$^{3}$}}
\newcommand{\affc}{{$^{2}$}}

\newcommand{\eqcont}{{$^{\dagger}$}}
\newcommand\tab[1][0.7cm]{\hspace*{#1}}

\usepackage[T1]{fontenc}

\usepackage[utf8]{inputenc}

\usepackage{microtype}

\usepackage{arydshln}

\usepackage{xspace}
\usepackage{listings} 
\newcommand{\babi}{{bAbI}\xspace}
\newcommand{\clutrr}{{CLUTRR}\xspace}
\newcommand{\brk}{\textbf{[B]}\xspace}
\newcommand{\brknospace}{\textbf{[B]}}
\newcommand{\brkname}{{BPT}\xspace}

\newcommand{\added}[1]{{#1}}

\title{Breakpoint Transformers for Modeling and Tracking Intermediate Beliefs}

\author{Kyle Richardson\affc\eqcont \tab Ronen Tamari\affa\eqcont\thanks{\tab[0.1cm] Work begun during an internship at the Allen Institute.} \tab Oren Sultan\affa \\ \quad {\bf Reut Tsarfaty}\affc\textsuperscript{,}\affb \tab {\bf Dafna Shahaf}\affa \tab {\bf Ashish Sabharwal}\affc \\
  \\
  \affa The Hebrew University of Jerusalem  \tab \affc Allen Institute for AI
   \tab \affb Bar-Ilan University \\
  {\small \texttt{\{ronent,orens,dshahaf\}@cs.huji.ac.il},  \texttt{\{kyler,reutt,ashishs\}@allenai.org}}
}
\begin{document}
\maketitle

\def\thefootnote{\eqcont}\footnotetext{Equal contribution.}\def\thefootnote{\arabic{footnote}}

\begin{abstract}

Can we teach natural language understanding models to track their beliefs through intermediate points in text? We propose a representation learning framework called breakpoint modeling that allows for learning of this type. Given any text encoder and data marked with intermediate states (\emph{breakpoints}) along with corresponding textual queries viewed as true/false propositions (i.e., the candidate \emph{beliefs} of a model, consisting of information changing through time) our approach trains models in an efficient and end-to-end fashion to build intermediate representations that facilitate teaching and direct querying of beliefs at arbitrary points alongside solving other end tasks. To show the benefit of our approach, we experiment with a diverse set of NLU tasks including \added{relational} reasoning on \clutrr and narrative understanding on bAbI. Using novel belief prediction tasks for both tasks, we show the benefit of our main \emph{breakpoint transformer}, based on T5, over conventional representation learning approaches in terms of processing efficiency, prediction accuracy and prediction consistency, all with minimal to no effect on corresponding QA end-tasks.  To show the feasibility of incorporating our belief tracker into more complex reasoning pipelines, we also obtain SOTA performance on the three-tiered reasoning challenge for the TRIP benchmark (around 23-32\% absolute improvement on Tasks 2-3).\footnote{\added{Project code available at \url{https://github.com/allenai/situation_modeling}}.} 

\end{abstract}

\section{Introduction}

\begin{figure}
    \centering
    \includegraphics[scale=0.46]{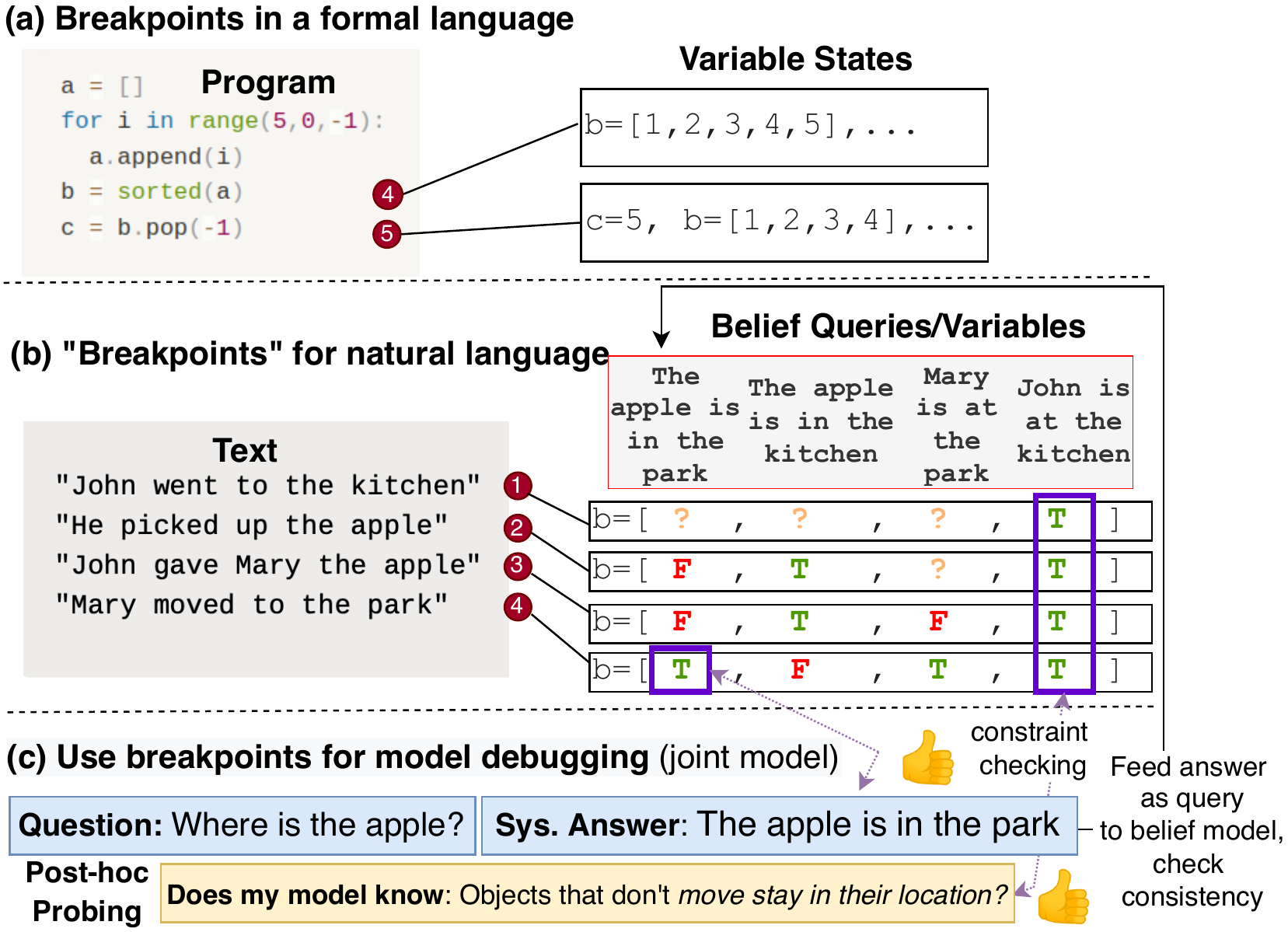} 
         
    \caption{Deep narrative understanding in natural language (bottom) involves the ability to answer \textbf{queries} about arbitrary intermediate points in a given story. We liken this task to \emph{breakpoints} in programming (top), or reporting the state of a program at different stages of execution, facilitating human inspection of model beliefs and consistency with end-task behavior (bottom).}
    \label{fig:story}
\end{figure}
Despite considerable progress made recently in natural language understanding (NLU), driven largely by advances in language model pre-training \cite{devlin2018bert,raffel2019exploring}  and the development of large-scale NLU benchmarks \cite{wang2018glue}, understanding the behavior of models remains a formidable and highly consequential challenge for model safety. Such a challenge is particularly acute in tasks such as narrative understanding, where one must piece together many individual (possibly implicit) facts through time in order to solve problems. For example, in the story in Figure~\ref{fig:story}, answering the question \emph{Where is the apple?} requires knowing how to track objects through time (e.g., knowing the location of the \emph{John} and \emph{Mary} and their interaction) and how to compartmentalize other types of knowledge across the story. In such a setting, where models are trained to narrowly answer questions, a natural question arises: \emph{do models acquire the kind of requisite background knowledge and world tracking abilities, and ultimately learn representations that give rise to correct \emph{beliefs}\footnote{\label{footnote:belief} Similar in spirit to \citet{kassner2021beliefbank}, we define a \emph{belief} as an attribution of a truth value to a proposition relative to a context or \emph{partial information state} \cite{landman2012structures}. E.g., a belief that \emph{John is in the kitchen} is $\mathbf{true}$ in the context immediately following the event \emph{John went to the kitchen}.} about intermediate states?}

A chief difficulty in answering such questions is that directly inspecting the propositional attitudes  of our current models remains a formidable challenge due to the latent nature of their knowledge. Such a complication also makes it unclear what the right interface should be for eliciting beliefs in the first place (e.g., how can we determine if a model believes a proposition \emph{John is in the kitchen} at an arbitrary point in text?).  In addition, for tasks such as QA, story contexts and questions are usually encoded jointly (often with full attention over context and query), which makes it difficult to tease apart a model's understanding of a story independent of each question. Entangled story and question representations can be inefficient when scaling to a large space of questions, particularly for novel combinations of questions and stories~\citep{tamari2021dyna}. Such entangled representations also allow models to exploit spurious patterns in questions that inflate performance \cite{kaushik2018much} and hinder interpretability.

\begin{figure}
    \centering
    \includegraphics[scale=0.55]{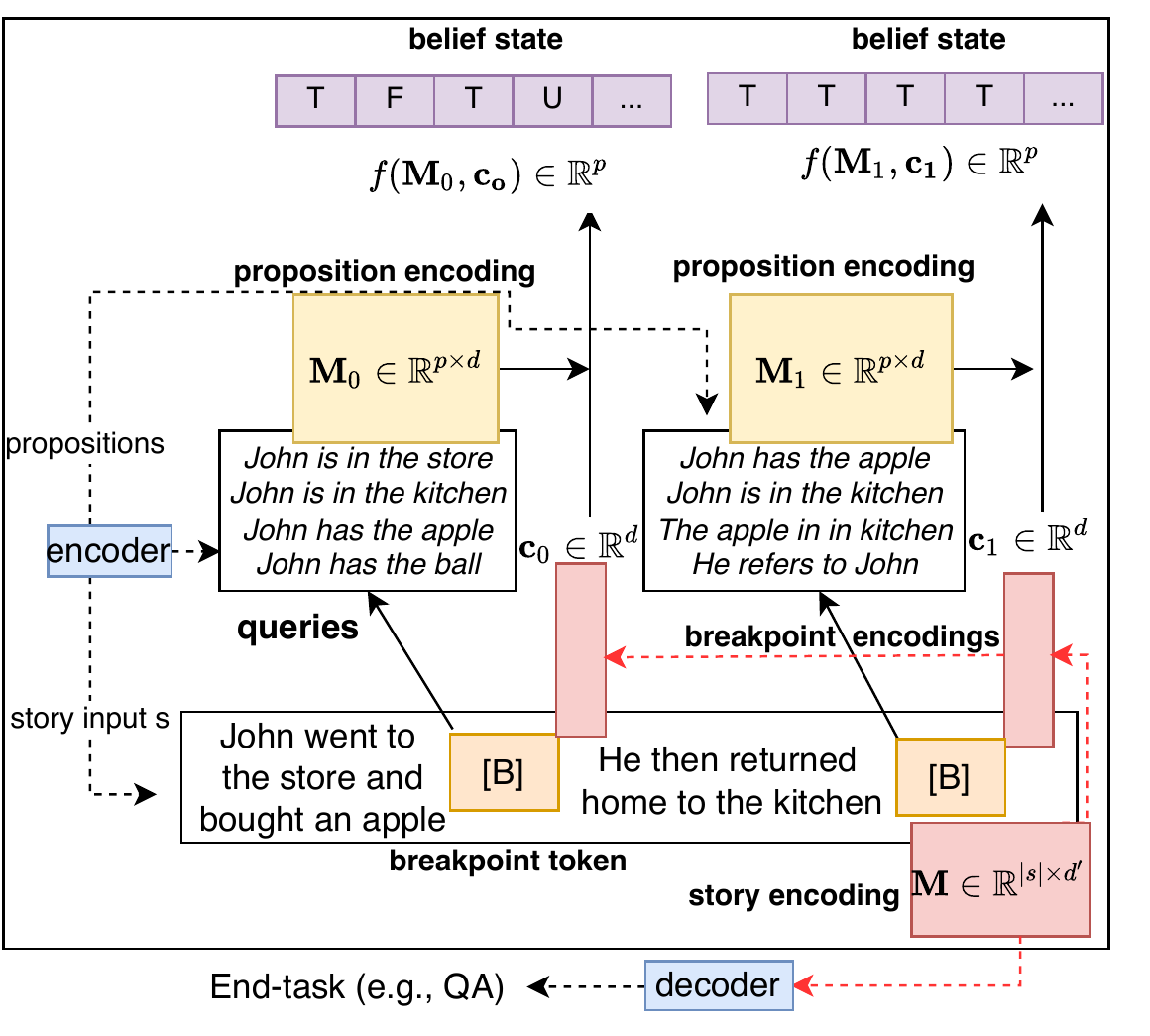}
    \caption{A high-level view of our modeling approach. For a given story and a set of textual \textbf{queries} corresponding to intermediate points in the story (\textbf{breakpoints}), truth assignments are assigned to queries to form \textbf{belief states} based on a projection over encodings of breakpoints and individual \textbf{proposition encoding}s using a single task-specific \textbf{encoder}.}
    \label{fig:breakpoint_arch}
\end{figure}

\begin{figure*}[ht]

	\centering 
	{\scriptsize
	\begin{tabular}{| p{1.7cm} p{4.7cm} p{8.2cm} |}
	    \hline 
		\multicolumn{1}{|c}{\textbf{Task}} & \multicolumn{1}{c}{\textbf{Example Stories}} & \multicolumn{1}{c|}{\textbf{Breakpoint Propositions}} \\ \hline  
		\textbf{Relational \newline Reasoning}  \newline (\clutrr) & John is the brother of Susan \hlgray{\textbf{[B]}$_{1}$} Susan's mother is Janice \hlgray{\textbf{[B]}$_{2}$}, ...  & 
		    \hlgray{$\mathbf{P}_{1}$} : \{ `Susan is the sister of John' \textcolor{green!45!black}{$\mathbf{true}$}, `Susan is the sister-in-law of Janice` \textcolor{red}{$\mathbf{false}$},`Janice is the mother of John` \textcolor{orange}{unk} \} \newline \hlgray{$\mathbf{P}_{2}$} : \{ `Janice is the mother of John` \textcolor{green!45!black}{$\mathbf{true}$}, `John is the father of Janice` \textcolor{red}{$\mathbf{false}$}, ...\}  \\ \hline 
		\textbf{Story \newline Understanding} \newline (\babi) & John moved to the kitchen \hlgray{\textbf{[B]}$_{1}$} He picked up an apple \hlgray{\textbf{[B]}$_{2}$} John then gave the apple to Mary \hlgray{\textbf{[B]$_{3}$}} ... &
		    \hlgray{$\mathbf{P}_{1}$} :\{ `John has the apple` \textcolor{red}{$\mathbf{false}$}, `John is in the kitchen` \textcolor{green!45!black}{$\mathbf{true}$},...\} \newline \hlgray{$\mathbf{P}_{2}$} : \{ `John has the apple` \textcolor{green!45!black}{$\mathbf{true}$},`John is in the kitchen`...\}... \newline \hlgray{$\mathbf{P}_{3}$} : \{ `John has the apple` \textcolor{red}{$\mathbf{false}$}, `Mary has the apple`  \textcolor{green!45!black}{$\mathbf{true}$} \}   \\ \hline 
		\textbf{Commonsense}  \newline (TRIP) & Tom dropped his radio ...carpet. \hlgray{\textbf{[B]}$_{1}$} The radio broke .. \hlgray{\textbf{[B]}$_{2}$} Tom turned on the radio ... \hlgray{\textbf{[B]$_{3}$}} ... &
		    \hlgray{$\mathbf{P}_{2}$} : \{ `radio is in pieces` \textcolor{green!45!black}{$\mathbf{true}$}, `radio is powered` \textcolor{red}{$\mathbf{false}$}\}, ...  \newline \hlgray{$\mathbf{P}_{3}$} : \{ `radio was powered` \textcolor{green!45!black}{$\mathbf{true}$} \}
		    \\ \hline  
	\end{tabular}}

\caption{Three tasks rendered as \textbf{stories} with special breakpoint tokens \hlgray{$\brk_{j}$} (for convenience, marked with an index $j$). Each intermediate breakpoint is aligned to a set of propositions \hlgray{$\mathbf{P}_{j}$} marked with truth conditions (i.e., $\textcolor{green!45!black}{\mathbf{true}},\textcolor{red}{\mathbf{false}}$, \textcolor{orange}{unknown}) corresponding to the truth value of each proposition at that breakpoint.}
\label{fig:breakpoints}
\end{figure*}


We present a model-agnostic representation learning framework called \emph{breakpoint modeling} that facilitates teaching models to have propositional beliefs at arbitrary points in stories (or \emph{breakpoints}) using ordinary textual queries as our interface language. Our general modeling approach is illustrated in Figure~\ref{fig:breakpoint_arch}. Given any task-specific encoder and data marked with the intermediate state of interest (or \emph{breakpoints}, denoted throughout as $\brk$) along with a set of textual queries (i.e., the candidate \emph{beliefs} provided in training as auxiliary intermediate supervision), models are trained in an end-to-end fashion to learn intermediate task-specific representations (pooled from single encodings of stories) that jointly facilitate making correct and consistent belief predictions efficiently across a large space of queries. Making an analogy with \emph{breakpoints} in programming (see top of Figure~\ref{fig:story}), we aim to \emph{simulate} stopping execution at intermediate points during a story to inspect the model's belief state (e.g., checking that a model's answers for QA are consistent with their beliefs and satisfy certain high-level constraints), as well as teach the model to have certain beliefs learned through intermediate supervision at training time.


%

Using a state-of-the-art pretrained model, T5 \cite{raffel2019exploring}, we develop and investigate a \emph{breakpoint transformer} to do belief prediction on three categories of tasks: \emph{narrative understanding} on \babi \cite{weston2015towards,tamari2021dyna}, \emph{relational reasoning} on \clutrr~\cite{sinha2019clutrr} and \emph{physical commonsense reasoning} over human authored stories on TRIP \cite{storks2021tiered}. In the former two cases, we focus on training and evaluating models on a novel belief prediction task. We report improvements over a conventional transformer-based representation learning approach \citep{reimers2019sentence} both in terms of prediction accuracy (4\% to 8\% absolute improvement on \clutrr dev) and belief consistency, all with significantly improved processing efficiency (i.e., minimal forward calls to the full transformer) and minimal effect on end-task performance when jointly trained with QA. In the latter case for TRIP, we show how to integrate our modeling approach into a more complex transformer pipeline and report state-of-the-art results on the three-tiered reasoning task (with 23-32\% absolute improvement on two component tasks) over existing task-specific architectures.





Taken together, our results show the viability of building an end-to-end trainable belief tracking mechanism and integrating it within existing transformer-based reasoning systems. To our knowledge, our work is among the first to look at at general-purpose sentence representation learning for intermediate states in text as a way to facilitate complex situation reasoning.

\section{Related Work}

Our work brings together two recent areas that aim to understand model behavior (broadly \emph{model probing}): probing of the type that includes finding neural correlates of high-level behavioral phenomena, modular structure in networks \cite{tenney2019bert,hewitt2019structural} on the one hand, as well as diagnostic testing, which aims to understand model competence through controlled input-output testing \cite{lake2018generalization,richardson2020probing}, or post-hoc consistency analysis \cite{kassner2021beliefbank}. Our work is more closely related to \citet{li2021implicit}, who show that partial world state information can be decoded from NLMs even without explicit supervision. In that work, state information is roughly localized to entity mentions, but varies across different datasets. Differently from such probing work, our breakpoint models are trained in a supervised manner to localize particular propositional information at particular locations (similar to \citet{geiger2021inducing}).


Our breakpoint model closely relates to \emph{late-interaction encoder} architectures that tease apart the encoding of problems and solutions. This includes the sentence transformer from \citet{reimers2019sentence}, which we compare against in our experiments, as well as \emph{read once transformers} \cite{lin2020readonce}, colBERT \cite{khattab2020colbert} and others. Given that the types of narrative tasks we focus on require modeling many intermediate points, we follow this work in putting an emphasis on representation and encoding efficiency. In contrast to this, and other related work on sentence representation learning \cite{gao2021simcse,ni2021sentence}, we uniquely focus on learning representations of intermediate states in text for complex situational reasoning.

We are also inspired by the situation modeling literature in cognitive science~\citep{GoldenRumelhart1993,Frank2003MW,Venhuizen2019}, and proposals for their integration with NLP research~\citep{tamari-etal-2020-language}. These works also studied neural models of narrative comprehension in carefully controlled micro-worlds, but typically focused on relatively short sentence-level inputs.

Our work also relates to efforts on building interpretable models by making the underlying reasoning processes  transparent, either through explicit decomposition \cite{andreas2016learning,khot2020text,bostrom2022natural} or generation of rationales \cite{camburu2018snli,wiegreffe2021teach} and other reasoning structures \cite{tafjord2020proofwriter,dalvi2021explaining,gontier2020measuring}.  In contrast, we focus on belief representations that are ultimately \emph{faithful} \cite{jacovi2020towards} to end-tasks by training knowledge directly into a model's task-specific representations.

\section{Breakpoint Modeling}
\label{sec:breakpoint_modeling}

The goal of breakpoint modeling is to capture the intermediate states and beliefs of models at arbitrary positions in text. Our models take \emph{stories} as inputs, or pieces of text containing one or more intermediate positions (\emph{breakpoints}), as well as sets of text \emph{propositions} that align to certain intermediate points (see Figure~\ref{fig:breakpoints}). Such propositions play the role of auxiliary supervision if provided at training time or as queries to the model for performing probing; when coupled with predictions they constitute the \emph{beliefs} of the model.

While breakpoint models can technically take different forms, their basic function is to assign encodings to  \added{intermediate states in text and their corresponding propositions} (\S~\ref{sec:encoding}) and to make predictions about the truth/falsity of each proposition (\S~\ref{sec:scoring}). Learning (\S~\ref{sec:learning}) reduces to the problem of teaching a model to have a correct and consistent set of beliefs for each target task given a set of representative intermediate propositions and beliefs provided at training time (\S~\ref{sec:sampling}).

\subsection{Breakpoint and Proposition Encoding}
\label{sec:encoding}

As illustrated in Figure~\ref{fig:breakpoints}, \textbf{stories} are texts consisting of $n$ tokens within which there can exist $m \geq 1$ arbitrarily selected intermediate points or \textbf{breakpoints}. For convenience, we will render a story $s$ in the following way: $s \texttt{:= }w_{1,b_{1}} \dots w_{\cdot,b_1} \brk \dots w_{\cdot,b_j} \dots \brk \dots w_{n,b_{m}} \brk$
where \brk is a special token used to explicitly mark position of each breakpoint $b_{j}$. \added{Intuitively, a breakpoint token represents all of the information in the story relevant to building an accurate belief state at the corresponding (intermediate) point in the text.} Associated with each $b_{j}$ is a set of text \textbf{propositions} $\mathbf{P}_{j} = \{ p_{1}, p_{2}, ... , p_{t} \}$. \added{Truth assignments to these text propositions constitute} the candidate beliefs at breakpoint $b_j$ \added{(in the sense of Footnote~\ref{footnote:belief})}.
\begin{figure}[t]
    \centering
    \includegraphics[scale=.47]{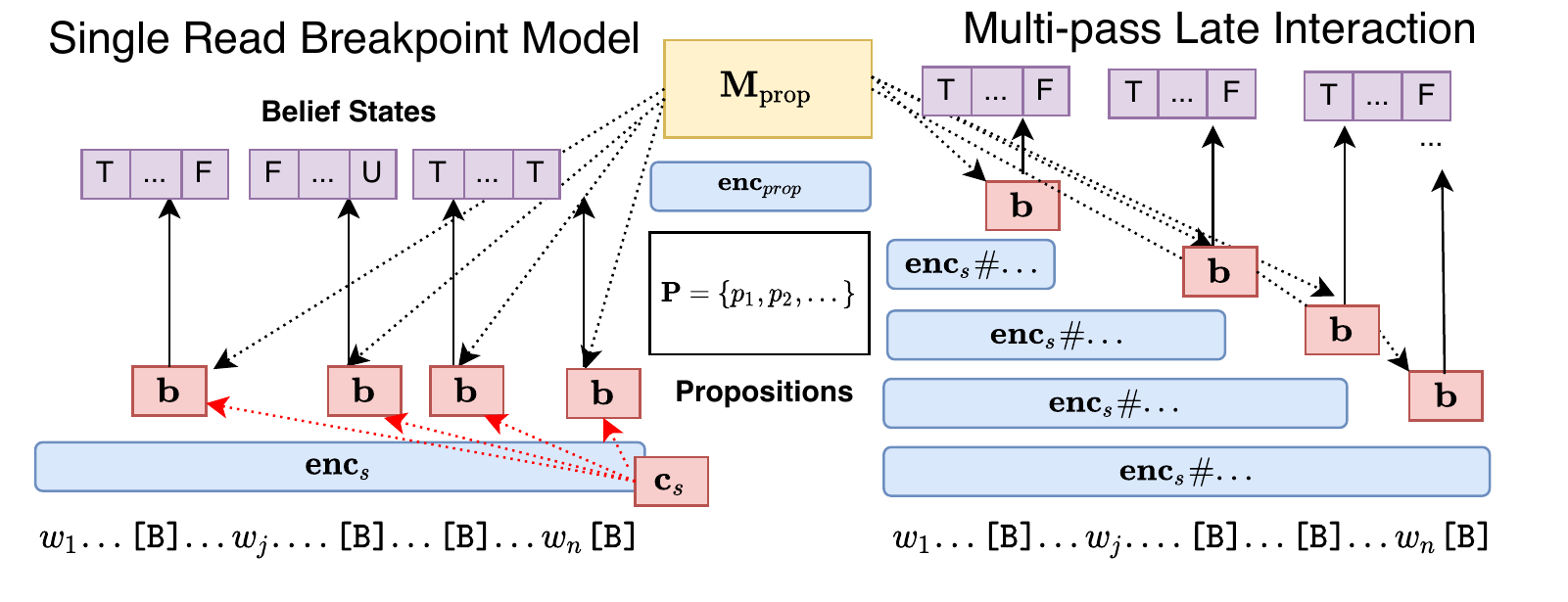}
    \caption{What is the best way to model intermediate states and beliefs with existing encoder models? An illustration of two late-interaction architectures we investigate (\textbf{Single Read} model, our main model described in \S~\ref{sec:breakpoint_modeling} and a \textbf{Multi-pass Late Interaction} model)}
    \label{fig:model_diagrams}
\end{figure}

At the core of any breakpoint model are two encoders, $\mathbf{enc}_{\text{story}}, \mathbf{enc}_{\text{prop}}$, that are used to generate a representation \added{or embedding} for each breakpoint in the story and each proposition, respectively. Representations of breakpoints $\mathbf{b} \in \mathbb{R}^{d}$  are pooled from a single encoding of an input story $s$: $\mathbf{c}_{s} \leftarrow \textbf{enc}_{\text{story}}(s) \in \mathbb{R}^{| s | \times d}$
and representations for propositions $\mathbf{c}_{\text{prop}} \in \mathbb{R}^{d}$ are obtained in a similar fashion using $\mathbf{enc}_{\text{prop}}$. While the choice of the encoder and the details of how pooling is done can vary (see details in \S\ref{sec:modeling}), in all of our models breakpoint representations $\mathbf{b}$ are obtained by taking projections of the hidden states of the \brk tokens from $\mathbf{c}_{s}$. We also investigate models that assume a \emph{siamese} architecture \cite{reimers2019sentence} where $\mathbf{enc}_{\text{story}}$ and $\mathbf{enc}_{\text{p}}$ are the same encoder.


An important property of breakpoint models is that all breakpoints representations $\mathbf{b}_j$  are obtained from a \emph{single read} encoding of each target story. We later compare this against a much less efficient approach that requires multiple forward passes through the story to obtain intermediate encodings (i.e., the \textbf{multi-pass} approach shown in Figure~\ref{fig:model_diagrams}).  Our model therefore stays within the spirit of a \emph{late-interaction} architecture \cite{khattab2020colbert} by using separate encodings of breakpoints and propositions, which allows us to scale to large sets of propositional queries.

\subsection{Proposition Scoring and Semantics}
\label{sec:scoring}

Given a breakpoint encoding $\mathbf{b}$ and an aligned proposition encoding $\mathbf{c}_{\text{prop}}$, a \textbf{proposition scorer} makes a prediction about a proposition at that breakpoint. As mentioned, our aim is to predict the truth \added{value} of a proposition at an intermediate state, which we take to be the model's \emph{belief} in that proposition. Our scorer takes the form of a classifier that maps a breakpoint encoding and proposition encoding to the discrete space $\{ \textcolor{black}{\mathbf{true}}, \textcolor{black}{\mathbf{false}}, \textcolor{black}{\mathbf{unknown}}\}$, following \citet{li2021implicit} and the annotation scheme from NLI \cite{dagan2005pascal,bowman2015large}.

To make clear that the interpretation of each proposition is tied to a specific breakpoint, we will use the symbolic notation from \citet{li2019logic} \added{and introduce three binary \emph{logical predicates} $\mathbf{E}, \mathbf{C},$ and $\mathbf{U}$. For each $b_j$ and $p \in \mathbf{P}_j$, these predicates capture whether $p$ is \textbf{entailed} by, is \textbf{contradicted} by, or has an \textbf{unknown} relation to the information in the text at breakpoint $b_{j}$, respectively. For instance, $\mathbf{E}(b_{j},p)$ is true if the text proposition $p$ is entailed by the story at breakpoint $b_j$.}

\subsection{Learning} 
\label{sec:learning}

\added{Suppose we have a dataset $D$ consisting of $n$ stories $\{ s^{(i)} \}_{i=1}^n$ along with the following additional information. For each story $s^{(i)}$, we have $m$ breakpoints $B^{(i)}$.\footnote{In general, $m$ depends on $i$. However, for simplicity of exposition, we use $m$ here instead of $m^{(i)}$.} For each such breakpoint $b_j$, we have $t$ labeled text propositions\footnote{Again, $t$ in general depends on both $i$ and $j$, but we use $t$ instead of $t^{(i)}_j$ here for simplicity.} $\mathbf{P}^{(i)}_j$, where each proposition $p_k \in \mathbf{P}^{(i)}_j$ is labeled with $y^{(i)}_{j,k} \in \{ \textcolor{black}{\mathbf{true}}, \textcolor{black}{\mathbf{false}}, \textcolor{black}{\mathbf{unknown}}\}$ indicating $p_k$'s truth value at breakpoint $b_j$. Using the above predicate logic notation, we can equivalently think of having, for each $p_k \in \mathbf{P}^{(i)}_j$, exactly one predicate $Y^{(i)}_{j,k} \in \{ \mathbf{E}, \mathbf{C}, \mathbf{U} \}$ annotated in $D$, with the semantics that $Y^{(i)}_{j,k}(b_j, p_k)$ is True (and the other two predicates for $b_j$ and $p_k$ are False).

The goal here is to learn a model that assigns truth values to all text propositions across all breakpoints---equivalently, truth values for all three logical predicates---in a way that maximally aligns with $D$. Semantically, this can be expressed as satisfying the logical formula \cite{li2019logic}:
\begin{align}
    {\footnotesize \bigwedge_{s^{(i)} \in D} \, \bigwedge_{b_{j} \in B^{(i)}} \, \bigwedge_{p_k \in \mathbf{P}_{j}^{(i)}} Y^{(i)}_{j,k}(b_{j},p_k)}
\label{eq:logic}
\end{align}
with the added constraint that for each story $s^{(i)}$ and all $j,k$, exactly one of $\mathbf{E}(b_j,p_k), \mathbf{C}(b_j,p_k),$ and $\mathbf{U}(b_j,p_k)$ is True.

Using $\Pr[y^{(i)}_{j,k}]$ to denote the model's probability corresponding to the predicate $\mathbf{Y}^{(i)}_{j,k}(b_j,p_k)$, this formula can be translated into the following loss using the \emph{product} translation from \citet{li2019logic}: 
\begin{align}
    \mathcal{L}_{\text{prop}} = \sum_{i=1}^{n} \sum_{j=1}^{m} \sum_{k=1}^t -\log \Pr[y^{(i)}_{j,k}]
\label{eq:prop_loss}
\end{align}
which yields the common cross-entropy loss that we use in our experiments.}

\subsection{Proposition Sampling}
\label{sec:sampling}

Propositions in breakpoint models have a dual role: when given at training time, they provide intermediate supervision for training models across different situation states. When given at inference time they allow for post-hoc probing of a model's beliefs. As shown in the Figure~\ref{fig:breakpoints}, propositions, in virtue of being ordinary text, can express many different types of information and thus provide an unbounded source of \emph{semantic supervision} \cite{hanjie2022semantic}, e.g., for expressing \emph{fluents}, or conditions that change through time in a story (e.g., \emph{John is in the kitchen}, or event pre/post-conditions (e.g., \emph{The radio \underline{was} powered} via English tense).

\begin{figure}
    \centering

\begin{tikzpicture}[
   level distance=20pt,
   sibling distance=40pt,
   every tree node/.style={anchor=north},
   every node/.append style={align=left}  
   scale=0.7, 
   every node/.style={scale=0.75}
]
\Tree 
    [.{\textcolor{green!45!black}{Qiana is Lisa's mother}}
        [.{\text{Derrick is}  \text{Lisa's father}}
        ]
        [.{\text{Qiana is}  \text{Derick's wife}}
        ]
    ]
\end{tikzpicture}

\vspace{.3cm}

{\scriptsize
\begin{tabular}{| p{1.7cm} | p{5cm} |}
\hline 
     \textbf{Story} & Lisa is Jerry's granddaughter \hlgray{\textbf{[B]}} Derrick is Lisa's father \hlgray{\textbf{[B]}} Qiana is Derrick's wife \hlgray{\textbf{[B]}} \\ \hline 
     \textbf{Atomic} \textbf{Belief} \newline \textbf{Annotations}: The basic facts that should be predicted  & 
     $\mathbf{E}$(1,`Jerry is the grandfather of Lisa`) \newline $\mathbf{E}$(1,`Derick is the father of Lisa`), \newline $\mathbf{C}$(1,`Lisa ..father of Derick`), \newline 
     $\mathbf{E}$(3,`Qiana..wife of Derick`),\newline $\mathbf{E}$(3,\textcolor{green!45!black}{Qiana is the mother of Lisa})  
     \\ \hline  
      \textbf{Knowledge}: Constraints to satisfy  &
      \texttt{(Implies}\texttt{(And}\textcolor{white}{dsd}$\mathbf{E}$(2,`Derick...father of Lisa`) \newline 
       \textcolor{white}{dsd}$\mathbf{E}$(3,`Qiana is the wife of Derick`)) \newline 
       $\mathbf{E}$(3,`Qiana is the mother of Lisa`)) \\ \hline 
      
\end{tabular}}

    \caption{How are intermediate propositions collected? An illustration of constructing intermediate propositions from \clutrr proof trees (above) in \citet{gontier2020measuring}. BOTTOM: An example ground constraint, which we use for analyzing consistency.}
    \label{fig:proof_example}
\end{figure}
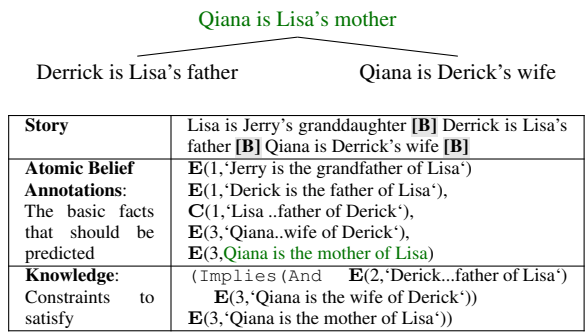

For training models to have beliefs, a necessary first step is to devise a \textbf{sampling policy} for generating these intermediate annotations. While such a strategy needs to be tailored to each target task, we experiment with a combination of extracting propositions from existing task annotations (Figure~\ref{fig:proof_example}) and generating propositions based on a set of \textbf{domain constraints} using the semantics of each target domain (details in the next section).

\section{Proposition Prediction Tasks} 
\label{sec:tasks}
We focus on three categories of tasks: text-based \textbf{relational reasoning}, \textbf{story understanding} and \textbf{commonsense reasoning}, each considered in turn. In the former two cases, we devise new proposition and belief prediction tasks that involve training on intermediate belief state annotations. We also include out-of-distribution (o.o.d) generalization tests beyond standard i.i.d (independent and identically distributed) evaluation. In the latter case, we recast an existing task in terms of breakpoint models to show the versatility of our approach in a more complex multi-task setting.  

\subsection{Relational Reasoning}
\clutrr \citet{sinha2019clutrr} focuses on QA over synthetic stories about family relations as shown in Figure~\ref{fig:breakpoints}, and has more recently been extended to focus on proof generation \cite{gontier2020measuring}. As illustrated in Figure~\ref{fig:proof_example}, we use the proof annotations in the latter work to generate intermediate propositions that track the time-course of family relations as they emerge at each new sentence.

Relying on the \emph{clean} subset of \clutrr stories \citet{sinha2019clutrr} and proof annotations, breakpoints are added after each sentence. Propositional renderings of the explicit story facts, as well as intermediate propositions revealed in the proof annotations, were then added to each corresponding breakpoint in the story and serve as the \emph{base} proposition set. From these base propositions, additional propositions, including negative and unknown propositions, were added using the following general constraints: \textbf{monotonicity}, that beliefs, once established to be true /false, cannot change; the \textbf{mutually exclusivity} of certain relations (e.g., \emph{X is the grandfather of Y} is mutually exclusive with \emph{X is the grandmother of Y}); \textbf{inverse relations} between certain relations (\emph{e.g.,} that \emph{$X_{fem}$ is a sister of $Y_{fem}$} means that $Y$ is a sister of $X$), and that all \textbf{non-deductively valid} propositions are unknown (i.e., with label $\mathbf{U}$).\footnote{\added{We note that all such constraints remain faithful to the semantics of the original tasks, such as \clutrr.}} Such \emph{ground} propositions constraints are included in the breakpoint annotations as symbolic expressions (see again Figure~\ref{fig:proof_example}) to allow for measuring model consistency at inference time (later in Figure~\ref{fig:training-size-ablation}. See details in \S~\ref{sec:clutrr-details}.


\noindent \textbf{o.o.d evaluation.}  Stories in \clutrr are characterized by their length $k$ (number of events) and \textbf{generalization} testing is usually performed to measure generalization. Our main datasets (later seen in Table~\ref{tab:cluttr_results}) consists of 13k training stories drawn from stories from $k=$2,..,5. We tune our models and evaluate on a mixture of in-domain and generalization stories of lengths $k$=2,..8 each containing around 1.5k stories (containing (avg) 10 propositions per breakpoint and 15 constraints per story). While these splits deviate from standard uses of \clutrr, we also compare against standard splits (i.e., training on $k=2,3$ and testing on $k'=2,..10$) to look at the ability of training joint belief prediction and QA models on the original \emph{QA task}.

\subsection{Story Understanding}
We experiment with the bAbI QA benchmark ~\citep{weston2015towards}, which contains questions over stories about agents in controlled micro-worlds (see Figure~\ref{fig:breakpoints}). As with \clutrr, the synthetic nature of domain makes it possible to automatically extract proposition annotations that express \emph{object location} (e.g., \emph{PersonX/ObjectX' is in Y}), \emph{object possession} (\emph{PersonX has ObjectY}), abstractions of \emph{event post-conditions} (e.g., \emph{PersonX took something} for the event \emph{PersonX grabbed the ball}) and \emph{pronoun} references (e.g., \emph{He refers to John}). We use the Dyna-bAbI task generator~\citep{tamari2021dyna} to generate initial base propositions and, similar with \clutrr, heuristically add more propositions using domain constraints (see \S~\ref{sec:babi-details} for more details).\footnote{In contrast to \clutrr and TRIP, bAbI does not have explicit \emph{unknown} proposition annotations, hence propositions either have label $\mathbf{E}$ or $\mathbf{C}$.} We use propositional versions of the 7-task set introduced in \citet{tamari2021dyna}. We specifically use the \emph{long-form} version of this set, where stories all contain 20 events/breakpoints, and train on 500 examples per task (totaling 3.5k+1.4k training/evaluation stories, with an average of 10 propositions per breakpoint and 123 constraints per story).

\noindent \textbf{o.o.d evaluation.} In addition to training and testing on this set, we also look at joint training on proposition prediction and the original QA task. For evaluation we also consider a more challenging \textbf{hardQA} generalization task from \citet{tamari2021dyna}, where the test set features compositions of concepts seen at training time. Appendix \ref{sec:babi-details} contains example inputs and further task details.

\subsection{Physical Commonsense Reasoning}
We apply our approach to the recently introduced Tiered Reasoning for Intuitive Physics (TRIP) dataset~\cite{storks2021tiered}. TRIP features a story plausibility end task, similar in scope to our proposition task, as well as a multi-tiered evaluation of models' reasoning process. Given a pair of highly similar human-authored short stories about everyday activities, models must jointly identify (1) the implausible story (\textbf{task1}) (2) a pair of conflicting sentences in the implausible story (\textbf{task2}) (3) the underlying physical states in those sentences causing the conflict (\textbf{task 3}). While \textbf{task3} takes the form of a breakpoint modeling task, where physical states are rendered as textual propositions, we model the first two tasks as text2text tasks using multi-task breakpoint models (details in the appendix and in \S~\ref{sec:modeling}). We use the original splits, consisting  of 675 plausible stories and 1472 implausible stories. \added{While we focus on the multi-tiered evaluation,  we devised a small \emph{filtered} dev set (644 stories) for later model analysis (Table~\ref{table:ablation}).}

\section{Modeling Details and Metrics}

Here we detail our main breakpoint transformer (\S\ref{sec:modeling}) following the framework in \S~\ref{sec:breakpoint_modeling} and all metrics used in our experiments (\S~\ref{sec:metrics}). 

\subsection{Modeling}
\label{sec:modeling}

\noindent \textbf{Encoder} We experimented with the T5 model \cite{raffel2019exploring} using the implementation from \citet{wolf2019huggingface}. T5's  bi-directional encoder was used for both our story encoder $\mathbf{enc}_{\text{story}}$ and proposition encoder $\mathbf{enc}_{\text{prop}}$.  While any comparable encoder would suffice, we chose T5 due its common use in NLU and ability to perform generation, which we used to implement other components in the multi-task models discussed below. \added{For efficiency reasons, we experimented with a combination of the smaller \textbf{T5-base} model (with 220M parameters) for datasets with long stories and many propositions (TRIP, \babi) and \textbf{T5-large} (with 770M parameters) for \clutrr.}




\noindent \textbf{Breakpoint and Proposition Embeddings} \added{For each story, individual breakpoint representations are first pooled from the \brk token hidden states in the story encodings $\mathbf{c}_{s}$ (see again Figure~\ref{fig:model_diagrams}). Following \citet{ni2021sentence}, a linear projection and L2 normalization is applied to each representation to construct initial breakpoint embeddings.  To allow for information transfer between different breakpoints, we then apply an additional self-attention layer (\textbf{sit-self}) over these resulting representations to obtain a \emph{self-attention} breakpoint representation (see \citet{fan2020addressing} for a similar idea), which gets concatenated with the initial representation to create the final breakpoint embedding. Operationally, the self-attention layer takes the form of a standard transformer block \cite{vaswani2017attention} with a single attention head. 

One subtlety in using a standard bi-directional encoder such as T5 is that each breakpoint token can look at future parts of the story. While the content of a breakpoint is often determined by the preceding sentence, in some cases it is important to have information about the future to obtain an accurate representation. For example, for the story \emph{John has the apple.$\brk_{1}$ He then moved to the kitchen $\brk_{2}$}, knowing that \emph{John} can't be \emph{in the kitchen} at $\brk_{1}$ (a \emph{pre-condition} of \emph{move} events) requires looking into the future. To limit the amount of future information in part of our breakpoint representations, however, future masking is applied in the breakpoint self-attention layer described above. 
}

To obtain a proposition embedding, we use the same T5 encoder over each text proposition prefixed with a special token, then take the hidden state of the target proposition.  A final proposition representation is then similarly obtained using the same linear projection and normalization layers. 




\noindent \textbf{Proposition Classifier} As in \citet{li2021implicit}, we use a bilinear layer for proposition classification (\text{score}$(\cdot)$). Using the notation from \S~\ref{sec:learning}, probabilities
\added{$\mathbf{\hat{y}}(\mathbf{b}_j,\mathbf{p}) = \langle \Pr[\mathbf{E}(b_j,p)], \Pr[\mathbf{C}(b_j,p)], \Pr[\mathbf{U}(b_j,p)] \rangle$ for the 3 truth values of a proposition $p$}
are computed in the following way using the final breakpoint representation $\mathbf{b}_j$ and proposition encoding $\mathbf{c}_{p}$:
\begin{align*}
    \text{score}(b_{j},p) &= \mathbf{b}_{j}^{T} \cdot \mathbf{M} \cdot \mathbf{c}_{p} + \mathbf{a}  \\ 
    \mathbf{\hat{y}}(b_{j},p) & = \mathbf{softmax}(\text{score}(b_{j},p)).
\end{align*}

\noindent \textbf{Learning} In addition to optimizing for the objective described in \S~\ref{sec:learning} ($\mathcal{L}_{\text{prop}}$), we also experiment with multi-task models trained to do generation ($\mathcal{L}_{gen}$) and QA ($\mathcal{L}_{qa}$), both of which are formulated as text2text tasks and optimized using standard cross-entropy-based training. In the former case, we investigate two analogues to the unsupervised \emph{denoising}  objectives from \cite{raffel2019exploring}, which aim to increase the amount of local information contained in breakpoint representations. 

The first is an \textbf{event generation} task that involves generating randomly chosen events from their right-most breakpoint encodings (e.g., generating the text \emph{Susan's mother is Janice} from the encoding of \brknospace$_{2}$ in Figure~\ref{fig:breakpoints}). The second, which is inspired by \citet{gontier2022does}, generates textual \textbf{abstractions} either of random events from breakpoints (in the case of TRIP, e.g., generating the abstracted text \emph{PERSON dropped his OBJ...} from \brknospace$_{1}$ in Figure~\ref{fig:breakpoints}) or random pairs of events in a story (e.g.,  generating the text \emph{A person received an apple} from the an encoding averaged from the two breakpoints \brknospace$_{2}$ and \brknospace$_{3}$ in Figure~\ref{fig:breakpoints}) (see additional details in \S~\ref{sec:sit-gen}).

Taken together, our full multi-task model's loss is: 
$\mathcal{L} = \lambda_{1}\mathcal{L}_{\text{prop}} + \lambda_{2} \mathcal{L}_{\text{qa}} + \lambda_{3}\mathcal{L}_{\text{gen}}$
where $\lambda_{\{{1,2,3}\}}$ are task weights manually tuned during training. We used ADAM as our optimizer \cite{kingma2014adam}. Standardly,  hyper-parameter tuning and model selection was performed via a random search search in the style of \citet{devlin2018bert} on held-out dev sets (see details in \S~\ref{sec:hyper}). \added{Unless stated otherwise, we report the average of three random restarts for all models and their standard deviations.}

\noindent \textbf{Baselines} We compare against two standard sentence representation learning approaches based on transformers and LSTMs. For the former we use the sentence transformer approach ~\citep{reimers2019sentence} applied to our task, and for the latter we use a model close to \citet{conneau2017supervised}. The set up is standard: stories and propositions are encoded separately using  a single encoder and collected via mean (transformer) and max (BILSTM) pooling then aggregated via concatenation (in the style of InferSent \citet{conneau2017supervised}) and fed into a softmax classifier to make a belief prediction. Importantly, these baselines models are much less efficient compared with our \emph{single read} breakpoint model, in that they require making multiple (\textbf{multi-pass late interaction}) forward passes through stories to create intermediate representations as  illustrated in Figure~\ref{fig:model_diagrams}. For the transformer models, with use the same T5 encoder as in the breakpoint models throughout all experiments.\footnote{As an additional check, we trained T5-based \emph{proposition-only} baseline, similar to the \emph{partial-input} baselines in NLI \cite{poliak2018hypothesis}, that make truth predictions from propositions \emph{alone} to check for spurious patterns. These always perform worse than our BILSTM baselines.}


Given that our breakpoint models take full story texts as input, to make the baselines fully comparable, we similarly feed in the full story on each read with a similar special token (\emph{\#}) to mark the target intermediate point (e.g., In the story \emph{John went to the store. He bought an apple} we feed the text \emph{John went to the store. \# He bought an apple} when modeling the first breakpoint).


\noindent \textbf{Joint Modeling} For \clutrr and bAbI, we also compare our multi-task breakpoint model trained for QA against T5 and Bart \cite{lewis2019bart}, both fine-tuned solely for QA.

\subsection{Metrics}
\label{sec:metrics}

For proposition prediction tasks we measure overall \textbf{proposition accuracy} (\%). Similarly for QA experiments, we follow other work in measuring exact match \textbf{EM accuracy} (\%) against a model's generated output. For some of our analysis on \clutrr (Figure~\ref{table:ablation}), we measure the consistency of belief prediction using the \textbf{global consistency} metric $\rho$ from  \citet{li2019logic}, which measures the fraction of stories containing one or more constraint violation using the constraint annotations described in \S~\ref{sec:tasks}. For example, using the constraint on the bottom Figure~\ref{fig:proof_example}, we first have the model make predictions about the constituent propositions (1. \emph{Derick is the father of Lisa}, 2. \emph{Qiana is the wife of Derick}. 3. etc..) and see if those predictions symbolically satisfy the constraint.

For TRIP, we follow exactly the 3-tiered evaluation of \citet{storks2021tiered}. We calculate: \textbf{Plausibility (task 1):} \% of instances where the implausible story was correctly identified. \textbf{Consistency (task 2):} \% of correctly identified implausible stories where the conflicting sentences were correctly identified. \textbf{Verifiability (task 3)}: \% of instances with correct plausibility/consistency predictions, where all relevant physical states are also identified.

\section{Results and Discussion} 
\label{sec:results}

We focus on the following questions: 1. \emph{Can our main model effectively and efficiently solve our new belief \textbf{proposition prediction} tasks (introduced in \S~\ref{sec:tasks}) and model intermediate state?} 2. \emph{Can we effectively integrate our breakpoint model into \textbf{joint models} for solving more complex tasks?} 


\begin{table}[ht]
    \centering
    {\scriptsize
    \begin{tabular}{| l  c| }
        \hline  
        \multicolumn{2}{|c|}{Proposition Prediction} \\ 
        \textbf{Model} &    Dev / Test Set + \textcolor{gray}{(std)} (Acc \%) \\ \hline   
         
         Majority Baseline  & 44.60 / 41.60 \phantom{\tiny ($\pm x.xx$)} \\
         BILSTM (Multi-pass) & 60.36  / 58.59 \textcolor{gray}{\tiny ($\pm 0.24$)} \\ 
         T5-large (Multi-pass)  & 81.41  / 81.94 \textcolor{gray}{\tiny ($\pm 0.17$)} \\ \cdashline{1-2} 
         \textbf{\brkname-large} & \textbf{85.16} / \textbf{85.24} \textcolor{gray}{\tiny ($\pm 0.34$)} \\ \hline 
    \end{tabular}}
    \\[.1cm]
    {\scriptsize
    \begin{tabular}{| p{1.4cm} p{2.5cm} p{2.5cm} |}
        \hline
        \multicolumn{3}{|c|}{Question-answering, \textbf{dev / test} + \textcolor{gray}{\text{(std)}}, (\textbf{EM} Acc \%)} \\ 
        \textbf{Model} & \multicolumn{1}{c}{i.i.d} & \multicolumn{1}{c|}{\textbf{generalization}}   \\ \hline
        FT-T5-base  &  99.00  / 99.78 \textcolor{gray}{\tiny ($\pm 0.19$)}  & 84.19  / \textbf{75.13} \textcolor{gray}{\tiny ($\pm0.94$)} \\ 
        FT-Bart-base & 98.65  / 98.94 \textcolor{gray}{\tiny ($\pm 0.78$)}  & 83.21   / 70.42 \textcolor{gray}{\tiny ($\pm 1.23$)}   \\ \cdashline{1-3}
        \textbf{\brkname-base} & \textbf{99.24}   / 99.75 \textcolor{gray}{\tiny ($\pm 0.19$)}  &  83.61  / 74.84 \textcolor{gray}{\tiny ($\pm 0.89$)} \\ \hline

    \end{tabular}}
    \caption{TOP: Proposition prediction results on \clutrr on the main mix dev and test sets comparing our breakpoint model (\textbf{BPT}) with baselines. BOTTOM: Evaluation on standard \clutrr QA ($k=2,3$) comparing our breakpoint model trained joint with QA  to fined-tuned (\textbf{FT}) T5 and Bart models.}
    \label{tab:cluttr_results}
\end{table}

\begin{figure}[ht]
    \centering
    \begin{tabular}{p{3.3cm} l}
    \includegraphics[scale=.23]{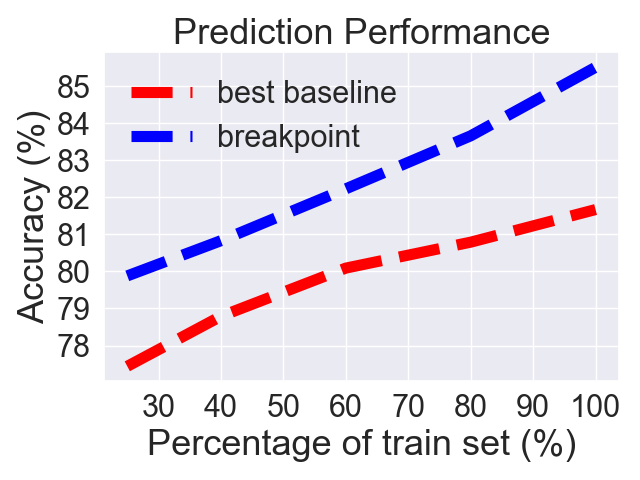} & \includegraphics[scale=.23]{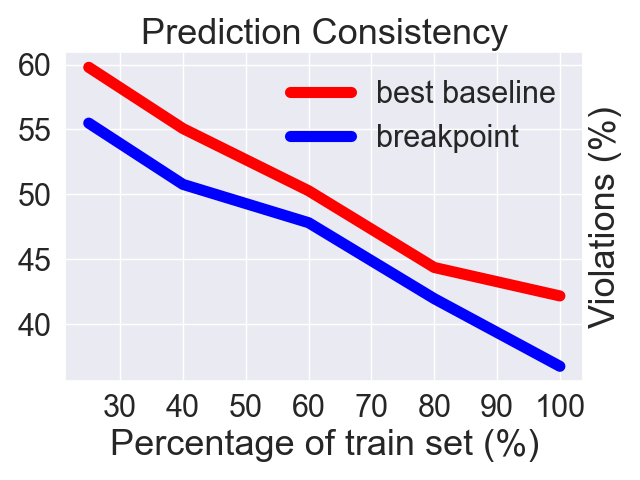}
    \end{tabular}
    \vspace{-.2cm}
    \caption{Effect of training data size on proposition prediction (left) and global consistency $\rho$ (right, \emph{lower is better}), on \clutrr dev (\added{best of 3 random runs}).}
    \label{fig:training-size-ablation}
\end{figure}

\noindent  \textbf{Proposition Prediction} We found breakpoint models to be effective at our proposition prediction tasks, most notably improving on the transformer \textbf{Multi-pass} baselines for \clutrr prediction from 81.9 to 85.2 (top of Table~\ref{tab:cluttr_results}, both an over 23\% improvement over our BILSTM baseline, suggesting task difficulty). Based on the plots in Figure~\ref{fig:training-size-ablation}, we also found our models to be more efficient learners (e.g., achieving comparable performance to baselines using only 60\% training data) and to exhibit less global constraint violations in the i.i.d setting (with a 6\% reduction in constraint violations $\rho$), thus leading to more consistent belief states. 

\begin{table}[ht]
    \centering
    {\scriptsize
     
        \begin{tabular}{| p{1.76cm}  p{3.2cm}  c |}
            \hline 
             \multicolumn{1}{|c}{\textbf{Model}} & \multicolumn{1}{c}{\textbf{i.i.d}} & \textbf{hard QA}  \\ \hline 
             & \begin{tabular}{p{1.3cm} | p{1.45cm}} \textbf{Prop.\%} & \textbf{QA\%} \end{tabular} & \textbf{QA\%} \\ \hline
             Majority & \begin{tabular}{p{1.3cm} | p{1.45cm}} 65.87 & -- \end{tabular}  & --  \\ \hline 
             \textbf{FT}-T5-base (QA) & \begin{tabular}{p{1.3cm} | p{1.45cm}} \textbf{--} & 97.29 {\tiny \textcolor{gray}{($\pm 0.14$)}}  \end{tabular} & 69.09 {\tiny \textcolor{gray}{($\pm 0.79$)}} \\ 
             \textbf{FT}-Bart-base & \begin{tabular}{p{1.3cm} | p{1.45cm}} \textbf{--} & \textbf{97.57} {\tiny \textcolor{gray}{($\pm 0.31$)}}  \end{tabular} & 67.21 {\tiny \textcolor{gray}{($\pm 0.80$)}} \\ \hline 
             \textbf{BILSTM} (Multi) &  \begin{tabular}{p{1.3cm} | p{1.45cm}} 80.2 {\tiny \textcolor{gray}{($\pm 0.16$)}} & --  \end{tabular} & -- \\ 
             \textbf{T5-base} (Multi) &  \begin{tabular}{p{1.3cm} | p{1.45cm}} \textbf{99.1} {\tiny \textcolor{gray}{($\pm 0.21$)}} & --  \end{tabular} &  -- \\ 
             \textbf{\brkname-base} & \begin{tabular}{p{1.3cm} | p{1.45cm}} 98.5 {\tiny \textcolor{gray}{($\pm 0.10$)}} & --  \end{tabular} & -- \\ \hline 
             \textbf{\brkname-base + QA} & \begin{tabular}{p{1.3cm} | p{1.45cm}} 98.5 {\tiny \textcolor{gray}{($\pm 0.10$)}} & 94.9 {\tiny \textcolor{gray}{($\pm 0.60$)}}  \end{tabular} & \textbf{70.51} {\tiny \textcolor{gray}{($\pm 0.29$)}} \\ \hline 
        \end{tabular}    
    }

    \caption{bAbI proposition prediction (\textbf{Prop. \%}) and \textbf{QA} performance on the main i.i.d and \textbf{hardQA} test sets.}
    \label{tab:babi_results}
\end{table}

For \babi (Table~\ref{tab:babi_results}) all transformer-based models achieve near perfect accuracy (and significantly outperform our BILSTM model); as such, models have near perfect consistency on the underlying constraints (not shown). Given that \babi stories are considerably longer than \clutrr stories (each containing 20 events/breakpoints), these results show the feasibility of modeling long contexts with our model and representing complex state information with individual breakpoints. In contrast to the baseline transformers, here we also see considerable practical improvements in \textbf{training time efficiency} due to our \emph{single read} architecture, resulting in a  54\% reduction in training time (from around 63 hours for multi-pass models to around 34 for ours on a single RTX A6000 GPU).

\begin{table}[ht]
    \centering
    {\scriptsize
    \begin{tabular}{| l | c c |}
        \hline 
         \multicolumn{3}{|c|}{\textbf{\clutrr} i.i.d vs. generalization (\textbf{gener.}) splits} \\ \hline 
         & \textbf{i.i.d}  ($k=2,..5$) & \textbf{gener.} \\ \hline
         \textbf{Baseline} (best run) & 94.54 \textcolor{gray}{($\rho=32.1$)} &  61.7 \textcolor{gray}{($\rho=96.2$)} \\ 
         \textbf{BPT} (best run) & 95.69 \textcolor{gray}{($\rho=25.9$)} & \textbf{69.2} \textcolor{gray}{($\rho=97.6$)}  \\ \hline  
    \end{tabular}}
    \caption{Comparison between i.i.d and compositional settings for \clutrr.}
    \label{table:compositional}
\end{table}

Our model's proposition prediction consistency is 7.5\% higher than that of the baseline, in terms of the $\rho$ metric reported in Table~\ref{table:compositional}. As an important caveat, however, in absolute terms, even our breakpoint model has much lower consistency on generalizations tasks (69.2\%) than in the i.i.d.\ setting (95.7\%). We discuss this further in \S~\ref{sec:limitations}.

\noindent \textbf{Joint Training} When trained jointly for both proposition prediction and QA, we found minimal to no impact on end-task performance, as shown on the bottom of Table~\ref{tab:cluttr_results} for \clutrr and in Table~\ref{tab:babi_results} for \babi (with a small improvement on the generalization QA task at the cost of a mere 2\% degradation in i.i.d.\ QA performance).
This shows the viability of integrating our belief tracking mechanism into existing transformer pipelines without significant performance drops. As first motivated in Figure~\ref{fig:story}, it also permits the development of more debuggable systems where the results of QA can be checked against the model's beliefs. 

\begingroup
\begin{table}[ht]
\centering 
{\scriptsize
\setlength{\tabcolsep}{4.5pt} 
\begin{tabular}{|cl l ll|}
\hline
\multicolumn{1}{|l}{\textbf{Split}}  & \textbf{Model} & \textbf{Task 1}\emph{(Plaus.)} & \textbf{Task 2}\emph{(Consist.)}  & \textbf{Task 3} \emph{(Verif.)} \\ \hline
\multirow{1}{*}{Dev}   & RoB        & 73.6            & 22.4            & 10.6            \\
&   \textbf{BPT-base}           & \textbf{81.99}{\tiny \textcolor{gray}{($\pm 0.91$)}}  & \textbf{58.07}{\tiny \textcolor{gray}{($\pm 0.76$)}}  & \textbf{36.44}{\tiny \textcolor{gray}{($\pm 0.53$)}}  \\ \cdashline{1-5}
\multirow{1}{*}{Test}  & RoB        & 72.9            & 19.1            & 9.1             \\
& \textbf{BPT-base}           & \textbf{80.55}{\tiny \textcolor{gray}{($\pm 1.20$)}}  & \textbf{53.83}{\tiny \textcolor{gray}{($\pm 1.65$)}}  & \textbf{32.37}{\tiny \textcolor{gray}{($\pm 0.27$)}}  \\ \hline
\end{tabular}}

\caption{Results on the TRIP 3-tiered physical commonsense reasoning benchmark, our main  \textbf{breakpoint} model (\textbf{\brkname}) compared against the RoBERTa-based approach (RoB) of \citet{storks2021tiered}.}
    \label{tab:trip_table}
\end{table}
\endgroup

Through our results on TRIP (Table~\ref{tab:trip_table}), we also see the viability of adding our belief tracking mechanism into more complex modeling pipelines. We were specifically able to obtain SOTA performance on this task and outperform the larger and highly tailored task-specific model architecture based on RoBERTa-large used by \citet{storks2021tiered}.


\begin{table}[ht]
    \centering
    {\scriptsize
    \begin{tabular}{| l | c c |}
         \hline
         \multicolumn{3}{|c |}{\textbf{\clutrr} (mix dev)} \\ \hline 
         & Prop. Acc\% & Global Violations $\rho$ \\
         \textbf{BPT-large} (best run) & 85.5 (\textcolor{red}{$\Delta$}) & 36.7 (\textcolor{red}{$\Delta$}) \\ \hline \cdashline{1-3} 
         \textcolor{gray}{- brk self-attn} & 77.3 \textcolor{red}{(-8.12)} & 54.3 \textcolor{red}{(-17.61)} \\  
         \textcolor{gray}{- event generation} & 82.1 \textcolor{red}{(-3.36)} & 41.8 \textcolor{red}{(-5.07)} \\ 
         \textcolor{gray}{- abstraction} & 82.1 \textcolor{red}{(-3.35)} & 42.8 \textcolor{red}{(-6.06)} \\ 
          \textcolor{gray}{BPT-base} & 81.8 \textcolor{red}{(-3.62)} & 44.3 \textcolor{red}{(-7.61)} \\ \hline
         \multicolumn{3}{|c|}{\textbf{TRIP} (fitered dev)} \\ \hline
         \textbf{BPT-base} (best run) & 92.8 (\textcolor{red}{$\Delta$}) & -- \\ \cdashline{1-3}
         \textcolor{gray}{- brk self-attn} & 89.43 \textcolor{red}{(- 3.36)} & -- \\  
         \textcolor{gray}{- event generation} & 89.43 \textcolor{red}{(- 3.36)} & -- \\  
         \textcolor{gray}{- abstraction} & 92.9 (+0.10) & --  \\ \hline 
    \end{tabular}}
    \caption{Breakpoint model feature ablations.}
    \label{table:ablation}
\end{table}

\noindent \textbf{Additional Analysis} We see in Table~\ref{table:ablation} for \clutrr that having an additional self-attention aggregation layer when constructing breakpoint representations (\textbf{-brk self-attn},  \S~\ref{sec:modeling}) is very important for accuracy and consistency (we find similar results for TRIP, bottom). This suggests that further improvements might be achieved through improved pooling and masking strategies for constructing breakpoint representations. We also see the advantages of having auxiliary generation losses (\textbf{event generation, abstraction}) for improving accuracy and performance. 


\section{Conclusion}



Being able to track the beliefs of models remains a formidable challenge at the forefront of model interpretability. In this paper, we presented a new representation learning framework, \emph{breakpoint modeling}, that facilitates end-to-end learning and tracking of beliefs at intermediate states in narrative text.  On a diverse set of NLU tasks, we show the benefit of our approach (based on T5) over conventional learning approaches in terms of improved belief prediction performance on new belief tracking tasks and processing efficiency. We also show the feasibility of recasting existing tasks into our framework and  integrating our approach into existing transformer-based NLU pipelines, which we believe can help to improve the interpretability of these models as part of this larger challenge.

\section*{Acknowledgements}
The authors thank the Aristo team for valuable feedback. This work was supported by the European Research Council (ERC) under the European Union's Horizon 2020 research and innovation programme (grant no. 852686, SIAM, Shahaf). Part of this research was also supported by the European Research Council, ERC-StG grant no.\ 677352 (Tsarfaty), which is gratefully acknowledged.

\section{Limitations}
\label{sec:limitations}

Below we summarize the main limitations of our current breakpoint models and the techniques pursued in this study. 

\paragraph{Compositional Generalization} Despite richer supervision over intermediate states, compositional generalization performance remains a significant challenge (on bAbI and \clutrr generalization splits, see \S\ref{sec:results}) for future work, which shows that our approach inherits many of the limitations in the generalization ability of large-scale LMs more broadly. Following \citet{kim-etal-2021-improving} and others, we hypothesize that the all-to-all attention employed by Transformers in creating token encodings (including the breakpoint tokens) is a factor in non-compositional behavior; such attention is more vulnerable to over-fitting spurious patterns. Accordingly, more advanced attention masking~\citep{kim-etal-2021-improving} and supervision~\citep{yin-etal-2021-compositional} approaches are promising directions to explore.

\paragraph{Our notion of ``belief''} While breakpoints provide an indication of intermediate model ``beliefs'', they are also different from beliefs in important ways. In particular, the causal relation between information represented in breakpoints and generated model outputs is unclear (see also \citet{li2021implicit} for similar caveats in standard NLMs). For example, models may generate outputs that are inconsistent with their own breakpoint belief states. Interestingly, breakpoint models may offer new ways to address these limitations by more explicitly representing intermediate reasoning steps; neural logic losses~\citep{li2019logic} can help enforce belief consistency between sets of propositions (\S\ref{sec:learning}). 

\paragraph{Task and domain limitations} Finally, our experiments are still limited to datasets involving relatively short (TRIP) and synthetic (bAbI, \clutrr) inputs with limited semantics. Further work is needed to address more natural and complex language to ultimately develop more robust breakpoint models. In contrast to standard end-to-end QA methods, breakpoint modeling requires more costly annotation, as training currently requires some form of supervision on intermediate states, beyond the final target output. Thus, developing new methods for collecting such annotations with minimal engineering effort remains a challenge.

\bibliography{anthology,custom}

\begin{thebibliography}{49}
\expandafter\ifx\csname natexlab\endcsname\relax\def\natexlab#1{#1}\fi

\bibitem[{Andreas et~al.(2016)Andreas, Rohrbach, Darrell, and
  Klein}]{andreas2016learning}
Jacob Andreas, Marcus Rohrbach, Trevor Darrell, and Dan Klein. 2016.
\newblock \href {https://doi.org/10.18653/v1/N16-1181} {Learning to compose
  neural networks for question answering}.
\newblock In \emph{Proceedings of the 2016 Conference of the North {A}merican
  Chapter of the Association for Computational Linguistics: Human Language
  Technologies}, pages 1545--1554, San Diego, California. Association for
  Computational Linguistics.

\bibitem[{Bostrom et~al.(2022)Bostrom, Sprague, Chaudhuri, and
  Durrett}]{bostrom2022natural}
Kaj Bostrom, Zayne Sprague, Swarat Chaudhuri, and Greg Durrett. 2022.
\newblock Natural language deduction through search over statement
  compositions.
\newblock \emph{arXiv preprint arXiv:2201.06028}.

\bibitem[{Bowman et~al.(2015)Bowman, Angeli, Potts, and
  Manning}]{bowman2015large}
Samuel~R. Bowman, Gabor Angeli, Christopher Potts, and Christopher~D. Manning.
  2015.
\newblock \href {https://doi.org/10.18653/v1/D15-1075} {A large annotated
  corpus for learning natural language inference}.
\newblock In \emph{Proceedings of the 2015 Conference on Empirical Methods in
  Natural Language Processing}, pages 632--642, Lisbon, Portugal. Association
  for Computational Linguistics.

\bibitem[{Camburu et~al.(2018)Camburu, Rockt{\"a}schel, Lukasiewicz, and
  Blunsom}]{camburu2018snli}
Oana-Maria Camburu, Tim Rockt{\"a}schel, Thomas Lukasiewicz, and Phil Blunsom.
  2018.
\newblock e-snli: Natural language inference with natural language
  explanations.
\newblock \emph{Advances in Neural Information Processing Systems}, 31.

\bibitem[{Conneau et~al.(2017)Conneau, Kiela, Schwenk, Barrault, and
  Bordes}]{conneau2017supervised}
Alexis Conneau, Douwe Kiela, Holger Schwenk, Lo{\"\i}c Barrault, and Antoine
  Bordes. 2017.
\newblock \href {https://doi.org/10.18653/v1/D17-1070} {Supervised learning of
  universal sentence representations from natural language inference data}.
\newblock In \emph{Proceedings of the 2017 Conference on Empirical Methods in
  Natural Language Processing}, pages 670--680, Copenhagen, Denmark.
  Association for Computational Linguistics.

\bibitem[{Dagan et~al.(2005)Dagan, Glickman, and Magnini}]{dagan2005pascal}
Ido Dagan, Oren Glickman, and Bernardo Magnini. 2005.
\newblock The pascal recognising textual entailment challenge.
\newblock In \emph{Machine Learning Challenges Workshop}, pages 177--190.
  Springer.

\bibitem[{Dalvi et~al.(2021)Dalvi, Jansen, Tafjord, Xie, Smith, Pipatanangkura,
  and Clark}]{dalvi2021explaining}
Bhavana Dalvi, Peter Jansen, Oyvind Tafjord, Zhengnan Xie, Hannah Smith,
  Leighanna Pipatanangkura, and Peter Clark. 2021.
\newblock \href {https://doi.org/10.18653/v1/2021.emnlp-main.585} {Explaining
  answers with entailment trees}.
\newblock In \emph{Proceedings of the 2021 Conference on Empirical Methods in
  Natural Language Processing}, pages 7358--7370, Online and Punta Cana,
  Dominican Republic. Association for Computational Linguistics.

\bibitem[{Devlin et~al.(2019)Devlin, Chang, Lee, and
  Toutanova}]{devlin2018bert}
Jacob Devlin, Ming-Wei Chang, Kenton Lee, and Kristina Toutanova. 2019.
\newblock \href {https://doi.org/10.18653/v1/N19-1423} {{BERT}: Pre-training of
  deep bidirectional transformers for language understanding}.
\newblock In \emph{Proceedings of the 2019 Conference of the North {A}merican
  Chapter of the Association for Computational Linguistics: Human Language
  Technologies, Volume 1 (Long and Short Papers)}, pages 4171--4186,
  Minneapolis, Minnesota. Association for Computational Linguistics.

\bibitem[{Fan et~al.(2020)Fan, Lavril, Grave, Joulin, and
  Sukhbaatar}]{fan2020addressing}
Angela Fan, Thibaut Lavril, Edouard Grave, Armand Joulin, and Sainbayar
  Sukhbaatar. 2020.
\newblock Addressing some limitations of transformers with feedback memory.
\newblock \emph{arXiv preprint arXiv:2002.09402}.

\bibitem[{Frank et~al.(2003)Frank, Koppen, Noordman, and Vonk}]{Frank2003MW}
Stefan~L. Frank, Mathieu Koppen, Leo~G.M. Noordman, and Wietske Vonk. 2003.
\newblock \href {https://doi.org/https://doi.org/10.1207/s15516709cog2706\_3}
  {Modeling knowledge-based inferences in story comprehension}.
\newblock \emph{Cognitive Science}, 27(6):875--910.

\bibitem[{Gao et~al.(2021)Gao, Yao, and Chen}]{gao2021simcse}
Tianyu Gao, Xingcheng Yao, and Danqi Chen. 2021.
\newblock \href {https://doi.org/10.18653/v1/2021.emnlp-main.552} {{S}im{CSE}:
  Simple contrastive learning of sentence embeddings}.
\newblock In \emph{Proceedings of the 2021 Conference on Empirical Methods in
  Natural Language Processing}, pages 6894--6910, Online and Punta Cana,
  Dominican Republic. Association for Computational Linguistics.

\bibitem[{Geiger et~al.(2021)Geiger, Wu, Lu, Rozner, Kreiss, Icard, Goodman,
  and Potts}]{geiger2021inducing}
Atticus Geiger, Zhengxuan Wu, Hanson Lu, Josh Rozner, Elisa Kreiss, Thomas
  Icard, Noah~D Goodman, and Christopher Potts. 2021.
\newblock Inducing causal structure for interpretable neural networks.
\newblock \emph{arXiv preprint arXiv:2112.00826}.

\bibitem[{Golden and Rumelhart(1993)}]{GoldenRumelhart1993}
Richard~M. Golden and David~E. Rumelhart. 1993.
\newblock \href {https://doi.org/10.1080/01638539309544839} {A parallel
  distributed processing model of story comprehension and recall}.
\newblock \emph{Discourse Processes}, 16(3):203--237.

\bibitem[{Gontier et~al.(2022)Gontier, Reddy, and Pal}]{gontier2022does}
Nicolas Gontier, Siva Reddy, and Christopher Pal. 2022.
\newblock Does entity abstraction help generative transformers reason?
\newblock \emph{arXiv preprint arXiv:2201.01787}.

\bibitem[{Gontier et~al.(2020)Gontier, Sinha, Reddy, and
  Pal}]{gontier2020measuring}
Nicolas Gontier, Koustuv Sinha, Siva Reddy, and Chris Pal. 2020.
\newblock \href
  {https://proceedings.neurips.cc/paper/2020/file/fc84ad56f9f547eb89c72b9bac209312-Paper.pdf}
  {Measuring systematic generalization in neural proof generation with
  transformers}.
\newblock In \emph{Advances in Neural Information Processing Systems},
  volume~33, page 22231–22242. Curran Associates, Inc.

\bibitem[{Haas(1987)}]{haas1987case}
Andrew~R Haas. 1987.
\newblock The case for domain-specific frame axioms.
\newblock In \emph{The Frame Problem in Artificial Intelligence}, pages
  343--348. Elsevier.

\bibitem[{Hanjie et~al.(2022)Hanjie, Deshpande, and
  Narasimhan}]{hanjie2022semantic}
Austin~W Hanjie, Ameet Deshpande, and Karthik Narasimhan. 2022.
\newblock Semantic supervision: Enabling generalization over output spaces.
\newblock \emph{arXiv preprint arXiv:2202.13100}.

\bibitem[{Hewitt and Manning(2019)}]{hewitt2019structural}
John Hewitt and Christopher~D Manning. 2019.
\newblock A structural probe for finding syntax in word representations.
\newblock In \emph{Proceedings of the 2019 Conference of the North American
  Chapter of the Association for Computational Linguistics: Human Language
  Technologies, Volume 1 (Long and Short Papers)}, pages 4129--4138.

\bibitem[{Jacovi and Goldberg(2020)}]{jacovi2020towards}
Alon Jacovi and Yoav Goldberg. 2020.
\newblock \href {https://doi.org/10.18653/v1/2020.acl-main.386} {Towards
  faithfully interpretable {NLP} systems: How should we define and evaluate
  faithfulness?}
\newblock In \emph{Proceedings of the 58th Annual Meeting of the Association
  for Computational Linguistics}, pages 4198--4205, Online. Association for
  Computational Linguistics.

\bibitem[{Kassner et~al.(2021)Kassner, Tafjord, Sch{\"u}tze, and
  Clark}]{kassner2021beliefbank}
Nora Kassner, Oyvind Tafjord, Hinrich Sch{\"u}tze, and Peter Clark. 2021.
\newblock \href {https://doi.org/10.18653/v1/2021.emnlp-main.697}
  {{B}elief{B}ank: Adding memory to a pre-trained language model for a
  systematic notion of belief}.
\newblock In \emph{Proceedings of the 2021 Conference on Empirical Methods in
  Natural Language Processing}, pages 8849--8861, Online and Punta Cana,
  Dominican Republic. Association for Computational Linguistics.

\bibitem[{Kaushik and Lipton(2018)}]{kaushik2018much}
Divyansh Kaushik and Zachary~C. Lipton. 2018.
\newblock \href {https://doi.org/10.18653/v1/D18-1546} {How much reading does
  reading comprehension require? a critical investigation of popular
  benchmarks}.
\newblock In \emph{Proceedings of the 2018 Conference on Empirical Methods in
  Natural Language Processing}, pages 5010--5015, Brussels, Belgium.
  Association for Computational Linguistics.

\bibitem[{Khattab and Zaharia(2020)}]{khattab2020colbert}
Omar Khattab and Matei Zaharia. 2020.
\newblock Colbert: Efficient and effective passage search via contextualized
  late interaction over bert.
\newblock In \emph{Proceedings of the 43rd International ACM SIGIR conference
  on research and development in Information Retrieval}, pages 39--48.

\bibitem[{Khot et~al.(2021)Khot, Khashabi, Richardson, Clark, and
  Sabharwal}]{khot2020text}
Tushar Khot, Daniel Khashabi, Kyle Richardson, Peter Clark, and Ashish
  Sabharwal. 2021.
\newblock \href {https://doi.org/10.18653/v1/2021.naacl-main.99} {Text modular
  networks: Learning to decompose tasks in the language of existing models}.
\newblock In \emph{Proceedings of the 2021 Conference of the North American
  Chapter of the Association for Computational Linguistics: Human Language
  Technologies}, pages 1264--1279, Online. Association for Computational
  Linguistics.

\bibitem[{Kim et~al.(2021)Kim, Ravikumar, Ainslie, and
  Ontanon}]{kim-etal-2021-improving}
Juyong Kim, Pradeep Ravikumar, Joshua Ainslie, and Santiago Ontanon. 2021.
\newblock \href {https://doi.org/10.18653/v1/2021.acl-short.81} {Improving
  compositional generalization in classification tasks via structure
  annotations}.
\newblock In \emph{Proceedings of the 59th Annual Meeting of the Association
  for Computational Linguistics and the 11th International Joint Conference on
  Natural Language Processing (Volume 2: Short Papers)}, pages 637--645,
  Online. Association for Computational Linguistics.

\bibitem[{Kingma and Ba(2014)}]{kingma2014adam}
Diederik~P Kingma and Jimmy Ba. 2014.
\newblock Adam: A method for stochastic optimization.
\newblock \emph{arXiv preprint arXiv:1412.6980}.

\bibitem[{Lake and Baroni(2018)}]{lake2018generalization}
Brenden Lake and Marco Baroni. 2018.
\newblock Generalization without systematicity: On the compositional skills of
  sequence-to-sequence recurrent networks.
\newblock In \emph{International conference on machine learning}, pages
  2873--2882. PMLR.

\bibitem[{Landman(2012)}]{landman2012structures}
Fred Landman. 2012.
\newblock \emph{Structures for semantics}, volume~45.
\newblock Springer Science \& Business Media.

\bibitem[{Lewis et~al.(2020)Lewis, Liu, Goyal, Ghazvininejad, Mohamed, Levy,
  Stoyanov, and Zettlemoyer}]{lewis2019bart}
Mike Lewis, Yinhan Liu, Naman Goyal, Marjan Ghazvininejad, Abdelrahman Mohamed,
  Omer Levy, Veselin Stoyanov, and Luke Zettlemoyer. 2020.
\newblock \href {https://doi.org/10.18653/v1/2020.acl-main.703} {{BART}:
  Denoising sequence-to-sequence pre-training for natural language generation,
  translation, and comprehension}.
\newblock In \emph{Proceedings of the 58th Annual Meeting of the Association
  for Computational Linguistics}, pages 7871--7880, Online. Association for
  Computational Linguistics.

\bibitem[{Li et~al.(2021)Li, Nye, and Andreas}]{li2021implicit}
Belinda~Z. Li, Maxwell Nye, and Jacob Andreas. 2021.
\newblock \href {https://doi.org/10.18653/v1/2021.acl-long.143} {Implicit
  representations of meaning in neural language models}.
\newblock In \emph{Proceedings of the 59th Annual Meeting of the Association
  for Computational Linguistics and the 11th International Joint Conference on
  Natural Language Processing (Volume 1: Long Papers)}, pages 1813--1827,
  Online. Association for Computational Linguistics.

\bibitem[{Li et~al.(2019)Li, Gupta, Mehta, and Srikumar}]{li2019logic}
Tao Li, Vivek Gupta, Maitrey Mehta, and Vivek Srikumar. 2019.
\newblock \href {https://doi.org/10.18653/v1/D19-1405} {A logic-driven
  framework for consistency of neural models}.
\newblock In \emph{Proceedings of the 2019 Conference on Empirical Methods in
  Natural Language Processing and the 9th International Joint Conference on
  Natural Language Processing (EMNLP-IJCNLP)}, pages 3924--3935, Hong Kong,
  China. Association for Computational Linguistics.

\bibitem[{Lin et~al.(2021)Lin, Sabharwal, and Khot}]{lin2020readonce}
Shih-Ting Lin, Ashish Sabharwal, and Tushar Khot. 2021.
\newblock \href {https://doi.org/10.18653/v1/2021.acl-long.554} {{R}ead{O}nce
  transformers: Reusable representations of text for transformers}.
\newblock In \emph{Proceedings of the 59th Annual Meeting of the Association
  for Computational Linguistics and the 11th International Joint Conference on
  Natural Language Processing (Volume 1: Long Papers)}, pages 7129--7141,
  Online. Association for Computational Linguistics.

\bibitem[{Ni et~al.(2022)Ni, Hernandez~Abrego, Constant, Ma, Hall, Cer, and
  Yang}]{ni2021sentence}
Jianmo Ni, Gustavo Hernandez~Abrego, Noah Constant, Ji~Ma, Keith Hall, Daniel
  Cer, and Yinfei Yang. 2022.
\newblock \href {https://doi.org/10.18653/v1/2022.findings-acl.146}
  {Sentence-t5: Scalable sentence encoders from pre-trained text-to-text
  models}.
\newblock In \emph{Findings of the Association for Computational Linguistics:
  ACL 2022}, pages 1864--1874, Dublin, Ireland. Association for Computational
  Linguistics.

\bibitem[{Poliak et~al.(2018)Poliak, Naradowsky, Haldar, Rudinger, and
  Van~Durme}]{poliak2018hypothesis}
Adam Poliak, Jason Naradowsky, Aparajita Haldar, Rachel Rudinger, and Benjamin
  Van~Durme. 2018.
\newblock \href {https://doi.org/10.18653/v1/S18-2023} {Hypothesis only
  baselines in natural language inference}.
\newblock In \emph{Proceedings of the Seventh Joint Conference on Lexical and
  Computational Semantics}, pages 180--191, New Orleans, Louisiana. Association
  for Computational Linguistics.

\bibitem[{Raffel et~al.(2020)Raffel, Shazeer, Roberts, Lee, Narang, Matena,
  Zhou, Li, and Liu}]{raffel2019exploring}
Colin Raffel, Noam Shazeer, Adam Roberts, Katherine Lee, Sharan Narang, Michael
  Matena, Yanqi Zhou, Wei Li, and Peter~J. Liu. 2020.
\newblock Exploring the limits of transfer learning with a unified text-to-text
  transformer.
\newblock \emph{J. Mach. Learn. Res.}, 21(1).

\bibitem[{Reimers and Gurevych(2019)}]{reimers2019sentence}
Nils Reimers and Iryna Gurevych. 2019.
\newblock \href {https://doi.org/10.18653/v1/D19-1410} {Sentence-{BERT}:
  Sentence embeddings using {S}iamese {BERT}-networks}.
\newblock In \emph{Proceedings of the 2019 Conference on Empirical Methods in
  Natural Language Processing and the 9th International Joint Conference on
  Natural Language Processing (EMNLP-IJCNLP)}, pages 3982--3992, Hong Kong,
  China. Association for Computational Linguistics.

\bibitem[{Richardson et~al.(2020)Richardson, Hu, Moss, and
  Sabharwal}]{richardson2020probing}
Kyle Richardson, Hai Hu, Lawrence Moss, and Ashish Sabharwal. 2020.
\newblock Probing natural language inference models through semantic fragments.
\newblock In \emph{Proceedings of the AAAI Conference on Artificial
  Intelligence}, pages 8713--8721.

\bibitem[{Sinha et~al.(2019)Sinha, Sodhani, Dong, Pineau, and
  Hamilton}]{sinha2019clutrr}
Koustuv Sinha, Shagun Sodhani, Jin Dong, Joelle Pineau, and William~L.
  Hamilton. 2019.
\newblock \href {https://doi.org/10.18653/v1/D19-1458} {{CLUTRR}: A diagnostic
  benchmark for inductive reasoning from text}.
\newblock In \emph{Proceedings of the 2019 Conference on Empirical Methods in
  Natural Language Processing and the 9th International Joint Conference on
  Natural Language Processing (EMNLP-IJCNLP)}, pages 4506--4515, Hong Kong,
  China. Association for Computational Linguistics.

\bibitem[{Storks et~al.(2021)Storks, Gao, Zhang, and Chai}]{storks2021tiered}
Shane Storks, Qiaozi Gao, Yichi Zhang, and Joyce Chai. 2021.
\newblock \href {https://doi.org/10.18653/v1/2021.findings-emnlp.422} {Tiered
  reasoning for intuitive physics: Toward verifiable commonsense language
  understanding}.
\newblock In \emph{Findings of the Association for Computational Linguistics:
  EMNLP 2021}, pages 4902--4918, Punta Cana, Dominican Republic. Association
  for Computational Linguistics.

\bibitem[{Tafjord et~al.(2021)Tafjord, Dalvi, and
  Clark}]{tafjord2020proofwriter}
Oyvind Tafjord, Bhavana Dalvi, and Peter Clark. 2021.
\newblock \href {https://doi.org/10.18653/v1/2021.findings-acl.317}
  {{P}roof{W}riter: Generating implications, proofs, and abductive statements
  over natural language}.
\newblock In \emph{Findings of the Association for Computational Linguistics:
  ACL-IJCNLP 2021}, pages 3621--3634, Online. Association for Computational
  Linguistics.

\bibitem[{Tamari et~al.(2022)Tamari, Richardson, Kahlon, Sar-shalom, Liu,
  Tsarfaty, and Shahaf}]{tamari2021dyna}
Ronen Tamari, Kyle Richardson, Noam Kahlon, Aviad Sar-shalom, Nelson~F. Liu,
  Reut Tsarfaty, and Dafna Shahaf. 2022.
\newblock \href {https://doi.org/10.18653/v1/2022.starsem-1.9}
  {{D}yna-b{A}b{I}: unlocking b{A}b{I}{'}s potential with dynamic synthetic
  benchmarking}.
\newblock In \emph{Proceedings of the 11th Joint Conference on Lexical and
  Computational Semantics}, pages 101--122, Seattle, Washington. Association
  for Computational Linguistics.

\bibitem[{Tamari et~al.(2020)Tamari, Shani, Hope, Petruck, Abend, and
  Shahaf}]{tamari-etal-2020-language}
Ronen Tamari, Chen Shani, Tom Hope, Miriam R~L Petruck, Omri Abend, and Dafna
  Shahaf. 2020.
\newblock \href {https://www.aclweb.org/anthology/2020.acl-main.559}
  {{L}anguage (re)modelling: {T}owards embodied language understanding}.
\newblock In \emph{Proceedings of the 58th Annual Meeting of the Association
  for Computational Linguistics}, pages 6268--6281, Online. Association for
  Computational Linguistics.

\bibitem[{Tenney et~al.(2019)Tenney, Das, and Pavlick}]{tenney2019bert}
Ian Tenney, Dipanjan Das, and Ellie Pavlick. 2019.
\newblock \href {https://doi.org/10.18653/v1/P19-1452} {{BERT} rediscovers the
  classical {NLP} pipeline}.
\newblock In \emph{Proceedings of the 57th Annual Meeting of the Association
  for Computational Linguistics}, pages 4593--4601, Florence, Italy.
  Association for Computational Linguistics.

\bibitem[{Vaswani et~al.(2017)Vaswani, Shazeer, Parmar, Uszkoreit, Jones,
  Gomez, Kaiser, and Polosukhin}]{vaswani2017attention}
Ashish Vaswani, Noam Shazeer, Niki Parmar, Jakob Uszkoreit, Llion Jones,
  Aidan~N Gomez, {\L}ukasz Kaiser, and Illia Polosukhin. 2017.
\newblock Attention is all you need.
\newblock In \emph{Advances in neural information processing systems}, pages
  5998--6008.

\bibitem[{Venhuizen et~al.(2019)Venhuizen, Crocker, and
  Brouwer}]{Venhuizen2019}
Noortje~J. Venhuizen, Matthew~W. Crocker, and Harm Brouwer. 2019.
\newblock \href {https://doi.org/10.1080/0163853X.2018.1448677}
  {Expectation-based comprehension: Modeling the interaction of world knowledge
  and linguistic experience}.
\newblock \emph{Discourse Processes}, 56(3):229--255.

\bibitem[{Wang et~al.(2018)Wang, Singh, Michael, Hill, Levy, and
  Bowman}]{wang2018glue}
Alex Wang, Amanpreet Singh, Julian Michael, Felix Hill, Omer Levy, and Samuel
  Bowman. 2018.
\newblock \href {https://doi.org/10.18653/v1/W18-5446} {{GLUE}: A multi-task
  benchmark and analysis platform for natural language understanding}.
\newblock In \emph{Proceedings of the 2018 {EMNLP} Workshop {B}lackbox{NLP}:
  Analyzing and Interpreting Neural Networks for {NLP}}, pages 353--355,
  Brussels, Belgium. Association for Computational Linguistics.

\bibitem[{Weston et~al.(2016)Weston, Bordes, Chopra, and
  Mikolov}]{weston2015towards}
Jason Weston, Antoine Bordes, Sumit Chopra, and Tom{\'{a}}s Mikolov. 2016.
\newblock Towards ai-complete question answering: {A} set of prerequisite toy
  tasks.
\newblock In \emph{4th International Conference on Learning Representations,
  {ICLR} 2016, San Juan, Puerto Rico, May 2-4, 2016, Conference Track
  Proceedings}.

\bibitem[{Wiegreffe and Marasovic(2021)}]{wiegreffe2021teach}
Sarah Wiegreffe and Ana Marasovic. 2021.
\newblock \href {https://openreview.net/forum?id=ogNcxJn32BZ} {Teach me to
  explain: A review of datasets for explainable natural language processing}.
\newblock In \emph{Thirty-fifth Conference on Neural Information Processing
  Systems Datasets and Benchmarks Track (Round 1)}.

\bibitem[{Wolf et~al.(2020)Wolf, Debut, Sanh, Chaumond, Delangue, Moi, Cistac,
  Rault, Louf, Funtowicz, Davison, Shleifer, von Platen, Ma, Jernite, Plu, Xu,
  Le~Scao, Gugger, Drame, Lhoest, and Rush}]{wolf2019huggingface}
Thomas Wolf, Lysandre Debut, Victor Sanh, Julien Chaumond, Clement Delangue,
  Anthony Moi, Pierric Cistac, Tim Rault, Remi Louf, Morgan Funtowicz, Joe
  Davison, Sam Shleifer, Patrick von Platen, Clara Ma, Yacine Jernite, Julien
  Plu, Canwen Xu, Teven Le~Scao, Sylvain Gugger, Mariama Drame, Quentin Lhoest,
  and Alexander Rush. 2020.
\newblock \href {https://doi.org/10.18653/v1/2020.emnlp-demos.6} {Transformers:
  State-of-the-art natural language processing}.
\newblock In \emph{Proceedings of the 2020 Conference on Empirical Methods in
  Natural Language Processing: System Demonstrations}, pages 38--45, Online.
  Association for Computational Linguistics.

\bibitem[{Yin et~al.(2021)Yin, Fang, Neubig, Pauls, Platanios, Su, Thomson, and
  Andreas}]{yin-etal-2021-compositional}
Pengcheng Yin, Hao Fang, Graham Neubig, Adam Pauls, Emmanouil~Antonios
  Platanios, Yu~Su, Sam Thomson, and Jacob Andreas. 2021.
\newblock \href {https://doi.org/10.18653/v1/2021.naacl-main.225}
  {Compositional generalization for neural semantic parsing via span-level
  supervised attention}.
\newblock In \emph{Proceedings of the 2021 Conference of the North American
  Chapter of the Association for Computational Linguistics: Human Language
  Technologies}, pages 2810--2823, Online. Association for Computational
  Linguistics.

\end{thebibliography}
\bibliographystyle{acl_natbib}

\clearpage
\appendix

\section{Dataset Details}

In this section, we provide additional details about all datasets. 

\subsection{TRIP}
\label{sec:trip-details}
As described in \S\ref{sec:tasks}, the TRIP benchmark consists of 3 tiered tasks: (1) plausibility (2) consistency (3) verifiability. To apply our model to TRIP, we convert the first two tasks to a text2text format: the first task involves taking two stories \texttt{(A) storyA (B) storyB \$plaus} as input text and producing a text label \{\texttt{A},\texttt{B}\} to identify the implausible story; task 2 involves taking a labeled story \texttt{sentence1 1 sentence2 2.. \$conflict} and generating the labels identifying the problematic sentences.\footnote{\texttt{\$plaus} and \texttt{\$conflict} are special tokens that prompt the model to output an answer for tasks 1 and 2, respectively.} We convert the third task to breakpoint format by converting state change labels to textual propositions associated with the corresponding timesteps. Figure \ref{fig:trip_instance} shows an example instance from the TRIP development set. Note that each task is effectively rendered as two instances: the first instance addresses task 1 (as QA), and the second jointly addresses tasks 2 (QA) and 3 (proposition prediction). 

State changes in TRIP are defined either as effects or preconditions~\citep{storks2021tiered} and this information must be preserved in the conversion to breakpoint format. Preconditions are propositions that hold \emph{before} a described event; for example, the proposition ``oven was open'' should be \textbf{true} before the sentence ``John closed the oven.'' Effect propositions are propositions that hold after a described event; the proposition ``oven is open'' should be \textbf{false} after ``John closed the oven.'' We represent precondition and effect propositions simply by modifying the proposition tense. Given breakpoint $b_{t}$, for  associated precondition propositions at time $t$, we use past tense (``oven was open''). For effect propositions at time $t$, we use present tense (``oven is open'').

While the TRIP data includes state information for all time steps and entities, we follow the official evaluation procedure\footnote{\url{https://github.com/sled-group/Verifiable-Coherent-NLU/blob/main/Verifiable-Coherent-NLU.ipynb}} and only score the subset of state changes defined to be relevant in the pair of conflict sentences. At training time, we use all available state change information for training. 

Finally, while most state changes in TRIP are attributes that can be \textbf{true}, \textbf{false} or \textbf{unknown} (and thus can be directly converted to proposition form), location attributes are formulated as $k$-way classification problems. For example, an object location attribute change is represented by 1 of 9 possible classes (see Table 5 in \citet{storks2021tiered} and blue propositions in Fig. \ref{fig:trip_instance}). To facilitate equivalent evaluation of $k$-class predictions with our breakpoint model, we consider the predicted \textbf{true} score for each of the possible $k$ propositions and take the maximum scoring proposition to be the predicted value.\footnote{Inspired by a similar method in \citet{li2021implicit}.}


\subsection{\babi}
\label{sec:babi-details}

\subsubsection{Proposition generation}
As detailed in \S~\ref{sec:tasks}, base propositions for \babi are generated using the Dyna-bAbI tool \cite{tamari2021dyna}. From this, new propositions are derived from the following general constraints: \textbf{location/possession uniqueness} that dictate that objects can only be in one location at a time and possessed by a single agent (e.g., \emph{John} cannot simultaneously be \emph{in the kitchen and living room}), \textbf{mutually exclusivity} between event types (e.g., that \emph{dropping a ball} is the opposite of \emph{picking up a ball}); \textbf{explanation frame rules} \cite{haas1987case} that dictate that objects, when left unchanged, maintain their location and their possession through time (e.g., \emph{John is in the kitchen} or \emph{John has the apple} stays true until there is an explicit event that changes this).

\subsubsection{Task details}
\begin{figure*}[t!]
\centering
\includegraphics[width=1\linewidth]{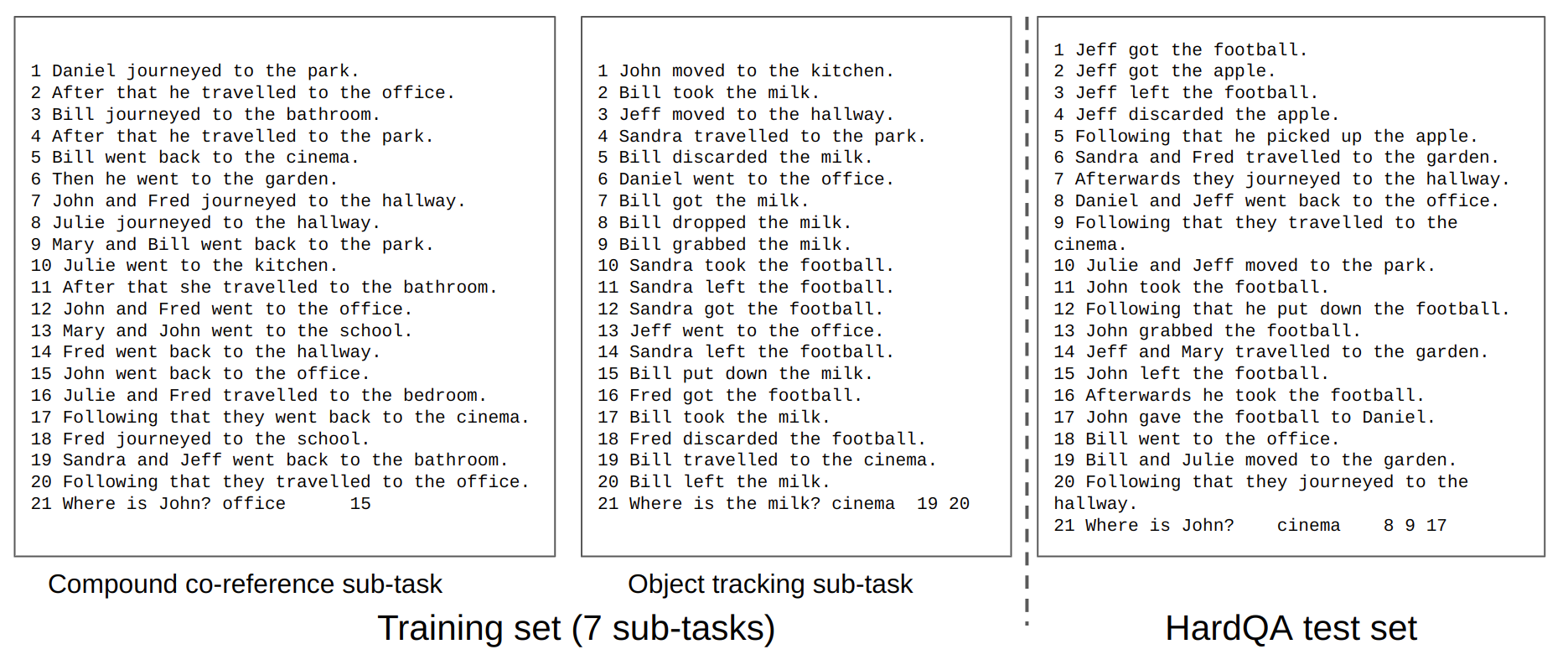}
\caption{\label{fig:babi_examples} Examples from the bAbI story understanding task. The train set includes 7 sub-tasks, such as co-reference and object tracking (left). The \textbf{hardQA} sample (right) incorporates novel compositions of concepts seen separately at training time. Beyond the question answering task, each example also includes proposition prediction at each time step (not shown here, see Figure \ref{fig:story}) for example.}
\end{figure*}

The training data includes 500 samples per task type, where the tasks follow the same structure as the \emph{concat(T7)} dataset described in \cite{tamari2021dyna} (Table 6 in that work), with the only difference being the story length which was fixed to 20 sentences to match the test data.
The \textbf{hardQA} generalization task was generated using the same settings as the \emph{mix(T7)} evaluation set from \cite{tamari2021dyna}, including the same 3 question types with 1,000 samples for each type (also Table 6 in \citet{tamari2021dyna}). Figure \ref{fig:babi_examples} shows example stories from the training and hardQA test splits.

\subsection{\clutrr}
\label{sec:clutrr-details}

We note that all of the underlying story data was generated from scratch and relies on the publicly available task generators from \citet{sinha2019clutrr} and \citet{gontier2020measuring}\footnote{See full details at: \url{https://github.com/facebookresearch/clutrr} and \url{https://github.com/NicolasAG/SGinPG}}. As detailed in \citet{gontier2020measuring}, leakage among the proofs and propositions in stories of the same $k$ can be a problem. Using some of their ideas, we avoided this by expanding the inventory of names used in training and abstracted names for parts of the training.  We verified the hardness of our data by training a no-story proposition-only baseline an found it to have low performance, and also manually verified all inference rules used for generating propositions. 

\begin{figure*}
\small

\colorlet{punct}{red!60!black}
\definecolor{background}{HTML}{EEEEEE}
\definecolor{delim}{RGB}{20,105,176}
\colorlet{numb}{magenta!60!black}
\lstdefinelanguage{json}{
    basicstyle=\normalfont\ttfamily,
    numbers=left,
    numberstyle=\scriptsize,
    stepnumber=1,
    numbersep=8pt,
    showstringspaces=false,
    breaklines=true,
    frame=lines,
    numbers=none,
    backgroundcolor=\color{background},
    literate=
     *{0}{{{\color{numb}0}}}{1}
      {1}{{{\color{numb}1}}}{1}
      {2}{{{\color{numb}2}}}{1}
      {3}{{{\color{numb}3}}}{1}
      {4}{{{\color{numb}4}}}{1}
      {5}{{{\color{numb}5}}}{1}
      {6}{{{\color{numb}6}}}{1}
      {7}{{{\color{numb}7}}}{1}
      {8}{{{\color{numb}8}}}{1}
      {9}{{{\color{numb}9}}}{1}
      {:}{{{\color{punct}{:}}}}{1}
      {,}{{{\color{punct}{,}}}}{1}
      {\{}{{{\color{delim}{\{}}}}{1}
      {\}}{{{\color{delim}{\}}}}}{1}
      {[}{{{\color{delim}{[}}}}{1}
      {]}{{{\color{delim}{]}}}}{1},
}

\def\bluecolor{\color{blue}}
\def\blackcolor{\color{black}}
\begin{lstlisting}[escapechar=@,language=json]
# Task 1 (plausibility)
{
	"example id": "414-C0-a",
	"question": "(A) John turned on the oven [B] John put the cake in the oven [B] John got the ice cream out [B] John put some ice cream in a red bowl [B] John put the red bowl in the oven [B] (B) John turned on the oven [B] John put the cake in the oven [B] John got the ice cream out [B] John put some ice cream in a red bowl [B] John put the rest of the ice cream in the fridge [B] $plaus",
	"answer": "B"
}

# Tasks 2 + 3 (consistency + verifiability)
{
	"example id": "414-C0-b",
	"question": "John turned on the oven 0 [B] John put the cake in the oven 1 [B] John got the ice cream out 2 [B] John put some ice cream in a red bowl 3 [B] John put the red bowl in the oven 4 [B] $conflict",
	"answer": "3,4"
	"proposition_lists": [
		[...], # sent. idx 0
		[...], # sent. idx 1
		[...], # sent. idx 2
		[
	  "red bowl is occupied", 
	  @\aftergroup\bluecolor@"ice cream is put into a container",
	  "ice cream does not move to a new location", 
	  "ice cream disappears",
	  "ice cream is picked up", 
	  "ice cream is put down", 
	  "ice cream is put on", "ice cream is removed", 
	  "ice cream is taken out of a container",
	  "ice cream moved somewhere new"@\aftergroup\blackcolor@,...
	        ], # sent. idx 3
		[
	"red bowl is put into a container", "oven was powered",
	"oven was open", "oven was turned on",...
	    ], # sent. idx 4
		],
        "labels": [
            [...], # sent. idx 0
            [...], # sent. idx 1
            [...], # sent. idx 2
            ["true", "true", "false", "false", "false", "false", "false", "false","false", "false",...], # sent. idx 3
            ["true", "true","true", "true",...], # sent. idx 4
	] 
}

\end{lstlisting}

\caption{Rendering of TRIP instance in breakpoint format. Breakpoint models can operate in standard text-to-text mode, generating output answers in response to questions, and additionally they can provide joint predictions over propositions associated with each sentence. Propositions in \textcolor{blue}{blue} indicate \textit{location} attributes which are evaluated as $k$-class predictions. See Appendix \ref{sec:trip-details} for further details on instance construction.}
\label{fig:trip_instance}
\end{figure*}

\section{Training details} 


\subsection{Hyper-parameters} 
\label{sec:hyper}

All hyper-parameter tuning for our main models was performed via a random search in the style of \citet{devlin2018bert}. Model selection was performed by selecting models with the highest validation accuracy for each task (e.g., proposition accuracy for our proposition tasks, exact match for the QA experiments). Unless noted otherwise, we report the average of models with the optimal hyper-parameters based on 3 random re-starts; early stopping was applied throughout. All experiments were performed on NVIDIA AX6000 GPU hardware on a single GPU.

\noindent \textbf{breakpoint models}: \emph{learning rate} (we experimented in the range of $\texttt{1e-3}$ to $\texttt{5e-6}$, we generally found $\texttt{5e-5}$ to be optimal for most experiments), \emph{number of epoch} (up to 35 for \clutrr, TRIP and 150 for \babi), \emph{batch size} (in the range of \emph{2} to \emph{16}, memory permitting, we found $2$ to be optimal for \babi and TRIP experiments, and $4$ for \clutrr) and \emph{weight decay} (set to $\texttt{0.001}$) and \emph{warmup steps} (from $\texttt{500}$ to $\texttt{1k}$ steps). See the project repository for further details

\noindent \textbf{joint models} For multi-task training, parameters $\lambda_{\{1,2,3\}}$ were hand tuned, with $\lambda_{1}$ set to \texttt{1.0} for all proposition prediction tasks (with $\lambda_{2}$=\texttt{0.1} for most tasks). For joint QA tasks, we found setting $\lambda_{1}=\texttt{1.0}$ and $\lambda_{1}=\texttt{1.0}$ to be optimal, with an initial warmup before turning on the proposition prediction loss (usually between $5$-$10$ epochs). Given the high cost of training the bAbI breakpoint QA model in Table~\ref{tab:babi_results}, the joint QA + prop models described on the last row start training from the \textbf{BPT-base} checkpoints described in the row above. 


\subsection{Auxiliary Generation Losses}
\label{sec:sit-gen}

As detailed in \S~\ref{sec:modeling}, we jointly trained our breakpoint models with additional generation losses that aim to mimic some of the unsupervised \emph{denoising objectives} used in \citet{raffel2019exploring}. Whereas in standard denoising you might try to generate from a text input \textbf{A dog <mask> while running} the output text \emph{<mask> barked loudly <mask>}, from an original text \emph{A dog barked loudly while running} (with full attention over the input text), in our case we try to generate from a story \textbf{John went to the store [B]$_{1}$ He then picked up the apple [B]$_{2}$} the raw event text \emph{John went to the store} from the corresponding raw breakpoint hidden state for the special token \textbf{[B]$_1$} alone. In addition to this \emph{event generation} task, we also experimented with a \emph{abstraction} generation task: given two stories in a batch and two random breakpoints within those stories, e.g., \textbf{John went to the kitchen [B]$_{1,1}$...} and \textbf{Mary went to the kitchen [B]$_{2,1}$..}, we ask the model to generate an abstract textual description of the two events only from the mean of the two breakpoint hidden states, i.e., \texttt{abstraction}$(\textbf{[B]}_{1,1},\textbf{[B]}_{2,1})=$ \emph{A person went to the kitchen}. (This was inspired by the abstraction generation ideas from \citet{gontier2022does}).  

During training, both forms of generation were done by randomly selecting a single breakpoint example and abstraction pair for each story in the batch and computing a standard loss over the generated texts and abstractions. Using symbolic annotations of both the \clutrr and \babi training events, a deterministic algorithm was implemented for creating abstracted texts on the fly for training. For TRIP, where logical annotations are not available, the abstraction task was replaced by the task of generating versions of text replaced with POS tags (e.g., \emph{John turned off the stove} would be turned into \emph{PER turned off the NOUN}).

\end{document}